\documentclass[runningheads]{llncs}

\usepackage{eccv}

\usepackage{eccvabbrv}

\usepackage{graphicx}
\usepackage{booktabs}
\usepackage[dvipsnames]{xcolor}

\usepackage{booktabs}
\usepackage{multirow}
\usepackage{animate}
\usepackage{media9}
\usepackage{graphicx}
\usepackage{float}
\usepackage{caption}
\usepackage{subcaption}
\usepackage{multicol}
\usepackage{todonotes}
\usepackage{wrapfig}
\usepackage{fancyhdr}

\DeclareMathOperator{\ml}{\mathcal{L}_{\text{ML}}}
\DeclareMathOperator{\mars}{\mathcal{L}_{\text{MARs}}}

\newcommand{\tcircle}[0]{\text{Circle}~\cite{circle}}
\newcommand{\tdrms}[0]{\text{DR-MS}~\cite{deen}}
\newcommand{\tntxent}[0]{\text{NTXent}~\cite{ntxent}}
\newcommand{\tpnp}[0]{\text{PNP}~\cite{pnp}}
\newcommand{\tproxyanchor}[0]{\text{Proxy Anchor}~\cite{proxyanchor}}
\newcommand{\tproxynca}[0]{\text{ProxyNCA++}~\cite{proxynca++}}
\newcommand{\tsubcenter}[0]{\text{Subcenter ArcFace}~\cite{subcenter}}
\newcommand{\tsupcon}[0]{\text{SupCon}~\cite{supcon}}
\newcommand{\tsynproxy}[0]{\text{Proxy Synthesis}~\cite{synproxy}}

\usepackage[accsupp]{axessibility}  % Improves PDF readability for those with disabilities.

\usepackage{hyperref}

\usepackage{orcidlink}

\begin{document}

% ---------------------------------------------------------------
% \title{MARs: Multi-view Attention Regularizations for Space Terrain Recognition}
\title{MARs: Multi-view Attention Regularizations for Patch-based Feature Recognition of Space Terrain}

% TODO REVIEW: If the paper title is too long for the running head, you can set
% an abbreviated paper title here. If not, comment out.
\titlerunning{MARs: Multi-view Attention Regularizations}

% TODO FINAL: Replace with your author list. 
% Include the authors' OCRID for the camera-ready version, if at all possible.
\author{Timothy Chase Jr\orcidlink{0009-0004-3075-0790} \and
Karthik Dantu\orcidlink{0000-0002-7497-6722}}

% TODO FINAL: Replace with an abbreviated list of authors.
\authorrunning{T.~Chase Jr, K.~Dantu}
% First names are abbreviated in the running head.
% If there are more than two authors, 'et al.' is used.

% TODO FINAL: Replace with your institution list.
\institute{University at Buffalo\\
\email{\{tbchase,kdantu\}@buffalo.edu}}

\maketitle

\begin{figure}[b!]
  \includegraphics[width=\textwidth]{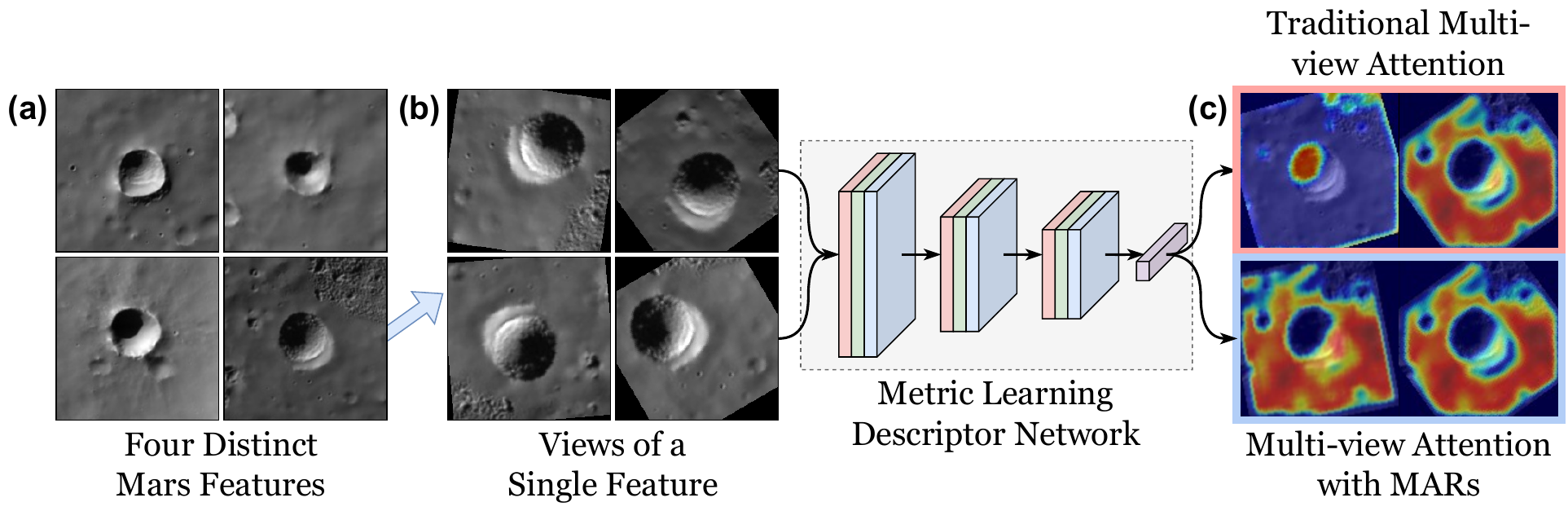}
  \caption{Patch-based features of space terrain exhibit extreme inter-class similarity and varying multi-view observations, which is difficult for metric learning to discern where attention focus is disparate. We propose Multi-view Attention Regularizations (MARs) to alleviate this issue and drive the attention of arbitrary viewpoints together.}
  \label{fig:intro}
\end{figure}

\begin{abstract}
The visual detection and tracking of surface terrain is required for spacecraft to safely land on or navigate within close proximity to celestial objects. Current approaches rely on template matching with pre-gathered patch-based features, which are expensive to obtain and a limiting factor in perceptual capability. While recent literature has focused on in-situ detection methods to enhance navigation and operational autonomy, robust description is still needed. In this work, we explore metric learning as the lightweight feature description mechanism and find that current solutions fail to address inter-class similarity and multi-view observational geometry. We attribute this to the view-unaware attention mechanism and introduce Multi-view Attention Regularizations (MARs) to constrain the channel and spatial attention across multiple feature views, regularizing the \textit{what} and \textit{where} of attention focus. We thoroughly analyze many modern metric learning losses with and without MARs and demonstrate improved terrain-feature recognition performance by upwards of 85\%. We additionally introduce the Luna-1 dataset, consisting of Moon crater landmarks and reference navigation frames from NASA mission data to support future research in this difficult task. Luna-1 and source code are publicly available at~\textcolor{magenta}{\url{https://droneslab.github.io/mars/}}.


\end{abstract}
\section{Introduction}\label{sec:intro}
Exploring deep space objects such as planets, comets, and asteroids involves ambitious and increasingly complex scientific pursuits. It has also been one of the earliest real-world applications of robotic autonomy. Advanced missions strive for spacecraft to land on or maneuver within close proximity to surfaces of highly irregular terrain and varying topography, which poses a significant challenge to spacecraft navigation as communication latency is often too great to permit any Earth-based assistance through radiometric tracking, real-time planning and control, or precise GPS positioning. More recently, these challenges are being addressed through the optical tracking of prominent surface terrain features to provide Terrain Relative Navigation (TRN). This approach has been validated on recent flagship missions including the landing of the Mars Perseverance Rover~\cite{m2020lvs} and the collection of asteroid regolith by OSIRIS-REx~\cite{nft}. With compute power limited by radiation-tolerant hardware, current approaches to TRN are template matching and correlation techniques using patch-based features (called landmarks) on static navigation maps that are collected and constructed \textit{a priori}~\cite{retina,m2020lvs,nft}. The set of landmarks and the underlying map require extensive pre-navigation costs and effort to obtain and develop. In the case of OSIRIS-REx, an estimated USD 68.5 million (roughly 25\% of the nine-year operations budget) was spent performing sufficient reconnaissance to gather and refine this data over a 1.5-year period~\cite{orex_cost, orex_phases}.

To reduce costs and accelerate mission timelines, it would be beneficial to detect and track these landmarks at navigation time similar to Simultaneous Localization and Mapping (SLAM) systems on Earth; a formulation that would also permit generalization to unseen and unexpected scenarios such as planetary weather~\cite{mars_dust} or asteroid ejection events~\cite{orex_eject}. SLAM is incredibly challenging to perform during TRN as space environments are generally unstructured, where low lighting and similarity in feature spaces create ambiguity and a lack of re-identifiability~\cite{slamprob1, slamprob2, slamprob3}. The use of learning-based solutions to overcome these challenges is possible with the recent inception of rad-hard inference accelerators~\cite{sclearn,myriad,scenic}, which enable in-situ terrain detection methods~\cite{yoco, det1, det2, det3, det4}.

Robust description of these detections remains an open problem. On Earth, this is similar to representation learning for tasks such as fine-grained classification, visual place recognition, and person re-identification. At the core of these applications is an objective to learn discriminative image embeddings for efficient similarity computation and downstream retrieval, which is commonly facilitated by metric learning. Compared to Earth-based recognition, however, landmark recognition in space is more nuanced where only one subclass of geological terrain is considered (e.g., \textit{crater}) with possibly thousands of individual instances to discern against (e.g., \textit{crater 570} vs \textit{crater 1181}). This is more fine-grained than even the most challenging of traditional benchmarks (e.g., CUB-200~\cite{cub}), demonstrated by~\autoref{fig:intro}-a. Apart from individual discernability, the terrain on the surface of planetary bodies, moons, and asteroids can vary widely in appearance from one observation to the next (\autoref{fig:intro}-b), which is difficult for metric learning to reason about on its own (\autoref{fig:intro}-c). 

In this work, we examine metric learning as it relates to landmark description during spacecraft TRN. We identify shortcomings in modern metric learning losses and consider poor performance an effect of the view-unaware attention mechanism included in modern architectures. We introduce Multi-view Attention Regularizations (MARs) to bolster recognition accuracy, training network attention to be implicitly view-aware and improving embedding distinguishability. Through additional similarity spaces, we constrain the \textit{what} and \textit{where} of attention information to enforce the consistency of focus between arbitrary feature views. Our approach is extensively validated on Earth, Mars, and Moon landmarks, where we introduce a photo-realistic dataset in the latter case. Experimental results demonstrate the effectiveness of our MARs learning constraint where attention between views is heavily correlated and recognition performance is greatly improved. Overall, we make the following contributions in this paper:
\begin{itemize}
    \item We study metric learning as the patch-based landmark descriptor for spacecraft navigation and perform extensive studies over traditional methods. We demonstrate shortcomings with terrestrial-based solutions and show correlations between the view-unaware attention mechanism and poor recognition performance in single-shot networks. To the best of our knowledge, this is the first study of its kind.
    \item We introduce Multi-view Attention Regularizations (MARs), a novel learning constraint to enforce the consistency of channel and spatial attention focus between arbitrary feature views.
    \item We release a new dataset, Luna-1, consisting of Moon crater landmarks and representative navigation frames using real-world NASA data, facilitating experimentation with multi-view and patch-based recognition systems in space navigation settings. 
    \item We demonstrate the utility of our MARs method, achieving state-of-the-art single-shot landmark description results on Earth, Mars, and Moon environments. Furthermore, we qualitatively showcase improved multi-view attention alignment using MARs.
\end{itemize}

\section{Related Work}\label{sec:relwork}
\underline{\textbf{Spacecraft Terrain Relative Navigation:}} Landmarks used for Terrain Relative Navigation (TRN) are collected \textit{a priori} through extensive surveying of the target body and crafted offline by human ground operators. RElative Terrain Imaging NAvigation (RETINA)~\cite{retina} and Natural Feature Tracking (NFT)~\cite{nft} are current asteroid-focused TRN methods that create 3D Digital Terrain Models (DTMs) by Stereophotoclinometry (SPC)~\cite{spc}. Visually prominent areas on the DTM are identified by hand, which are extracted as 2D image templates and uploaded to the spacecraft. Onboard, these templates (i.e., landmarks) are re-generated in SPC fashion to adjust shading based on the predicted illumination conditions of the surface. Navigation frames are then searched for correspondence by traditional image processing algorithms. The Mars Perseverance Landing Vision System (MP-LVS~\cite{m2020lvs}) deployed a similar technique during the landing phase of the Mars 2020 mission, which hand-picked landmarks on landing site survey imagery captured by other orbiting spacecraft at Mars.

Current TRN solutions include many shortcomings that are a detriment to mission cost, complexity, and time-to-science. The amount of pre-navigation imagery required is immense, and the subsequent time needed to hand-pick which ``features-to-track'' is extensive. The total number of features used by the system is incredibly sparse due to the level of human involvement, which drastically reduces perceptual capability and prevents reasoning over unseen areas. Onboard rendering of predicted landmark appearances severely limits frame rate, which can jeopardize spacecraft safety during critical phases of the mission. For example, the deployment of NFT on OSIRIS-REx executed at 0.0083 FPS, or one frame every two minutes, as it made contact with the surface~\cite{nft}.

\underline{\textbf{Terrestrial Recognition:}} The front-end vision in current TRN systems can be radically improved by leveraging rad-hard accelerators and object detection-style observation methods discussed in~\autoref{sec:intro}; although a robust description technique is required to close the loop. Earth-based tasks such as fine-grained visual classification (FGVC), visual place recognition (VPR), and person re-identification (Re-ID) intrinsically demonstrate this capability and reason over similar challenges, including high intra-class and low inter-class variances~\cite{earthrec1,earthrec2,earthrec3}, multi-view observations~\cite{earthrec4,earthrec5,earthrec6,earthrec8}, and appearance change over time~\cite{earthrec7,earthrec8}. Nevertheless, there are considerable challenges in adopting the current literature. Modern solutions to FGVC, VPR, and Re-ID are focused on description and retrieval problems at internet-scale~\cite{ilrsurv,ilrsurv2,cirsurv,fgsurv,vprsurv} and consequently have become more involved than a single-stage network. These methods employ multiple forward passes~\cite{mfp1,mfp2}, region proposals~\cite{mfp7,mfp8,mfp9,mfp10,mfp11}, model fusions~\cite{ff1,ff2,ff3,ff4,ff5,ff6,ff7,ff8}, multi-stage re-rankings~\cite{rr1,rr2,rr3}, and high-parameter transformer models~\cite{reranktrans, trans1,trans2,trans3,trans4,trans5,trans6,trans7,trans8,trans9,trans10}. As such, there is a primary concern about the physical execution of these techniques onboard resource-limited spaceflight computers~\cite{csp,sc3,sc3_mini}. Large models that cannot fit within accelerator caches must be executed in a hybrid manner, where model parameters are streamed from the host processor to the accelerator during inference. This has a detrimental effect on execution time~\cite{aero} and requires careful consideration, given that cache sizes in the current generation of spacecraft accelerators are small (e.g., 8 MB in~\cite{sclearn}).

Furthermore, TRN landmark recognition requires more granular reasoning than FGVC, VPR, and Re-ID, akin to frame-to-frame feature matching problems in SLAM. Recognition in FGVC, VPR, and Re-ID is performed by recalling instances from a pre-seeded database by global description~\cite{earthrec6,earthrec7}, where any viewpoint and domain generalization is generally a byproduct of learning with extremely large datasets~\cite{large1,earthrec7,large2} or the aggregation of large datasets~\cite{trans10}; a technique that is not currently adoptable due to the lack of space landmark datasets (two at the time of writing including the proposed Luna-1). Additionally, the \textit{sequential} nature of the TRN task needs consideration, 
% There is also a consideration of the \textit{sequential} nature of the TRN task 
where any recognition database is populated as samples are encountered instead of recalling against the entire population upfront (the effects of which have not been studied previously). Futhermore, the appearance differences between instances of a geological space-terrain feature (e.g., crater) are generally more subtle than traditional FGVC and Re-ID datasets making discernability more challenging~\cite{cub,stanfordcars, market}.

We identify metric learning as the core facilitator of discriminative representation learning in terrestrial tasks and examine its capabilities to permit lightweight, onboard-executable single-shot description networks for spacecraft TRN. Observing terrain features from a remote-sensing platform exhibits complex transformation spaces, however, which must be taken into consideration.

\underline{\textbf{Viewpoint Challenges and Attention:}} During TRN the observed target body is rotating and revolving distinctly from the spacecraft leading to an unconstrained appearance change in landmark illumination, translation, and rotation over time. Such a transformation space is generally uncommon in the literature (Re-ID would not expect a person observation to be upside-down for example~\cite{market}), and modern metric learning losses do not permit invariancy to these transformations directly. The convolutional layers used in modern networks are known to be equivariant to translations over the input image~\cite{eqsurv,eq1}, but are not naturally equivariant to rotations. Explicit in-network modifications for adding rotation equivariance have recently been explored including steerable filters~\cite{e2cnn,eq1}, multi-orientation feature extractions~\cite{eq5,eq6}, and alternative coordinate systems~\cite{eq2,eq3,eq4,ric}. Equivariant properties can be additionally learned~\cite{leq1,leq2}, which may be advantageous as a supplement to explicit mechanisms or when explicit mechanisms are themselves undesirable~\cite{leq3}. Learning equivariance is popular in the literature through batch-sampling, mining, and augmentation approaches~\cite{mleqsurv,contrastiveblog,mleq1,mleq2,mleq3,mleq4}. The remote sensing literature has studied similar techniques with the fusion of pre-trained group convolutions~\cite{rs1} and probabilistic formulations of metric space locations~\cite{ride}, although they are restrictive in their reasoning through trainings with pre-rotated data.

The explicit encoding of equivariant properties into the attention mechanism has recently been explored~\cite{attention2,attention3,attention4}. At large, however, analyzing \textit{learned} attention equivariance as it compares to these mechanisms (or the combination thereof) has not been studied previously. The Self-supervised Equivariant Attention Mechanism (SEAM)~\cite{attention1} is one of the only works that target attention-equivariance learning directly through self-supervised regularization. Multi-view attention similarity learning such as the Contrastive Attention Map Loss (CAML)~\cite{caml} has shown impressive equivariant properties as a byproduct of contrastive learning over attention maps. Although, integration of these methods within metric learning frameworks is a challenging task as SEAM requires Class Activation Maps (CAMs) and CAML targets foreground/background feature separation using image statistics from segmentation labels.

\section{Methodology}\label{sec:method}
Prior work demonstrates that attention has a large influence on recognition performance in multi-view settings, but the extent of this influence concerning equivariant properties (either encoded or learned) is unclear. Equivariance does not guarantee that attention, being a strictly learnable mechanism, will be identical between multiple views of the same feature; it only \textit{suggests} that it should be similar. An alignment of attention focus should lessen the downstream recognition difficulty, maximizing separability and view-dependent groupings in the embedding space, although such a constraint is not readily formulated in current multi-view metric learning pipelines. We suggest that any attention disagreement must be directly accounted for during the training, and propose a soft learning constraint to rectify any variance. This concept forms the basis of our proposed Multi-view Attention Regularizations (MARs), described in this section. We first introduce our learning framework and baseline network architecture in~\autoref{sec:learning} and~\autoref{sec:arch}. We then detail our constraint for aligning attention and frame the overall learning objective in~\autoref{sec:constraints}.

\subsection{Learning Framework}\label{sec:learning}
The framework for data augmentation and batch formation plays a critical role in multi-view similarity learning~\cite{mleq1,mleq2,mleq3,mleq4}, where we start by following the popular SimCLR~\cite{mleq1} method. SimCLR aims to maximize the learned representation similarity between augmented views of the same input. With training batch size $B$, we begin by sampling a minibatch of $B/2$ samples where each sample $x$ gets augmented by two distinct transformation operations to produce new views $x_1 = t_1(x)$ and $x_2 = t_2(x)$ where $t_1$ and $t_2$ are sampled from the same family of augmentations $\mathcal{T}$. $\mathcal{T}$ is a composition function of three image transformations that include a random brightness adjustment, rotation, and translation. An encoder network $f(\cdot)$ is applied to the augmented data to extract intermediate representations $h_1 = f(x_1)$ and $h_2 = f(x_2)$. These representations are in turn mapped to the metric space through projection head $g(\cdot)$ to yield embeddings $z_1 = g(h_1)$ and $z_2 = g(h_2)$. Given this $(z_1,z_2)$ positive pair, the other $2(B/2-1)$ embeddings in the minibatch are considered negative samples. The batch of $z$ embeddings is fed to any applicable metric learning loss $\mathcal{L}_{ML}$, as is the traditional metric learning process. To assist with the inter-class granularity of landmark recognition we additionally employ hard sample mining in traditional multi-similarity (MS)~\cite{multisim} fashion to yield $ap,p,an,n$ batch-indices where $ap,p$ represent anchor-positives and positives (simply the indices of the twice augmented images) and $an,n$ the indices of embeddings deemed similar by the MS metric but have different instance labels. 

\subsection{Network Architectures}\label{sec:arch}
With inspiration from large-scale Earth-based recognition networks~\cite{google_landmark,gl_first,mlsurvblog} we employ a ResNeXt-101~\cite{resnext} architecture with Squeeze-and-Excitation (SE)~\cite{se} attention as the baseline for encoder network $f(\cdot)$. Encoder $f(\cdot)$ is the primary bottleneck to onboard execution performance as it will hold the most parameters, and we select ResNeXt-101 as a middle ground between discriminative representation power and model size. Furthermore, we elect to stay on the larger end of model size in contrast to the ResNet-50 class~\cite{resnet50} to isolate representation power and examine the effects of different attention and equivariance setups. Our embedding projection head $g(\cdot)$ is a smaller network consisting of a Generalized Mean Pooling (GeM)~\cite{gem} layer followed by a linear (512), batch norm, and PReLU activation. In contrast to Earth-based recognition, we perform no functions other than a single shot $f(\cdot)$ and $g(\cdot)$ to descript instances.

\underline{\textbf{Encoding Rotational Equivariance:}} Augmentations applied in $\mathcal{T}$ mimic the unconstrained landmark appearance change found in spacecraft TRN (assuming the spacecraft is in a non-geosynchronous position relative to the target body). As the pose of the target body will be changing independently of the spacecraft we cannot assume rotated landmark views will be limited to anything less than a full 360 degree of change. Although data augmentation attempts to implicitly teach the network to be robust, reasoning over this level of extreme rotation remains a challenging property to learn. As such, we additionally seek to study the benefits of explicit rotational equivariance integration in $f(\cdot)$.

RIC-CNN~\cite{ric} develops a convolutional operation (the Rotation-Invariant Coordinate Convolution, RIC-C) based on a novel coordinate system that permits this equivariance as a replacement to standard convolutional layers. RIC-C extends the idea of deformable convolutions~\cite{deform} and does not require any transformation of the representation space of input images or intermediate features. This property enacts a simple and efficient implementation, which we leverage in this work by replacing all standard convolution operations in $f(\cdot)$ with RIC-C layers. For brevity, we refer interested readers to~\cite{ric} for the full account of the coordinate system and RIC-C operation.

\underline{\textbf{Spatial Attention:}} SE attention improves the interdependencies within feature maps by assigning weights to each channel and selecting the most relevant for a given input. This type of attention is focused on relevancy \textit{between} features alone (channel) and carries no understanding of relevancy \textit{within} an individual feature (spatial). We assume spatial attention has a critical role in multi-view metric learning and introduce this in $f(\cdot)$. Coordinate Attention (CA~\cite{ca}) provides spatial awareness through distinctive pooling operations in the height and width dimensions while preserving the channel dimensionality. This is in contrast to other spatial attention techniques such as the Convolutional Block Attention Module (CBAM~\cite{cbam}) that collapse channel information via pooling before learning spatial weight factors. We modify $f(\cdot)$ by replacing SE attention with CA.

\subsection{Forming Attention Similarity Constraints}\label{sec:constraints}
The inclusion of explicit rotational equivariant properties through RIC-C layers and the ability to attend spatially with CA is the basis for which we explore our proposed MARs constraint. During the training procedure, we seek to drive both the \textit{what} (channel) and the \textit{where} (spatial) elements focused by the attention mechanism together, without explicitly assuming that one view is correct in either of these aspects. This alignment is thus a moving target, where it is imperative to impose a soft constraint between them. In other words, it is undesirable to calculate a strict differentiation between attention maps at any point during training to avoid a collapse in attention information. The constraint should prioritize that the attention maps from each view evolve similarly over time. 

\underline{\textbf{Pose Normalization and Channel Reduction:}} To facilitate this\\constrained evolution we propose to introduce regularization terms by embedding attention into additional metric spaces. Let $A_i$ be the set of attention maps output from ResNeXt block $i \in N$ from $f(\cdot)$ where $N$ is the number of these blocks. For each positive pair in the training batch, we have multi-view attention maps $A_{i1}$ and $A_{i2}$. For each ResNeXt block outputting $A_{i}$, we output inverse transformation $t_i^{-1}$ where the translation parameters are adjusted relative to the spatial resolution of $A_i$. We apply the inverse transformation to normalize the translation and orientation (pose) of each attention map to equal that of the input image, yielding pose-normalized attention maps $\hat{A}_{i1} = t_{i1}^{-1}(A_{i1})$ and $\hat{A}_{i2} = t_{i2}^{-1}(A_{i2})$. To embed attention into additional metric spaces we employ mini variants of the projection head $g(\cdot)$, which do not include any linear layers for dimensionality reduction. Instead, we first reduce the channel dimension of $\hat{A}_{i} \in \mathbb{R}^{C \times H \times W}$ through a 1x1 convolution $\text{Conv}_{i}^{1}(\cdot)$ with reduction factor $r$ to yield $\hat{A}^r_{i} \in \mathbb{R}^{C/r \times H \times W}$. This process prevents obscurification, keeps the data correlated, and reduces learnable parameters. 

\underline{\textbf{Channel and Spatial Attention Embeddings:}} For positive and\\pose-normalized attention pairs $(\hat{A}^r_{i1},\hat{A}^r_{i2})$ we utilize the mini channel-wise ($c$) projection head $gc_i(\cdot)$ to produce channel attention embeddings $zc_{i1} = gc_i(\hat{A}^r_{i1})$ and $zc_{i2} = gc_i(\hat{A}^r_{i2})$. GeM pooling collapses the spatial dimensions to yield an embedding with length given by $C/r$ where $C$ is the channel dimension of the current ResNeXt block $i$. For spatial attention embeddings, we first perform height and width pooling (similar to CA) on $\hat{A}^r_{i}$. Specifically, given height ($y$) and width ($x$) pooling operators $\text{Ypool}(\cdot)$ and $\text{Xpool}(\cdot)$ we produce intermediate representations $hy_i = \text{Ypool}(\hat{A}^r_{i})$ and $hx_i = \text{Xpool}(\hat{A}^r_{i})$. These representations are input to mini spatial projection heads $gy_i(\cdot)$ and $gx_i(\cdot)$ to yield height and width embeddings $zy_{i} = gy_i(hy_{i})$ and $zx_{i} = gx_i(hx_{i})$. The mini projection heads $gc_i(\cdot)$, $gy_i(\cdot)$, and $gx_i(\cdot)$ do not share any parameters and are instantiated once per block $i \in N$. This allows distinct regularization on attention maps with the same channel-spatial resolution as well as calculating accurate batch-norm statistics that are channel, spatial-height, and spatial-width disparate.

\begin{figure*}
\centering
  \includegraphics[width=\textwidth]{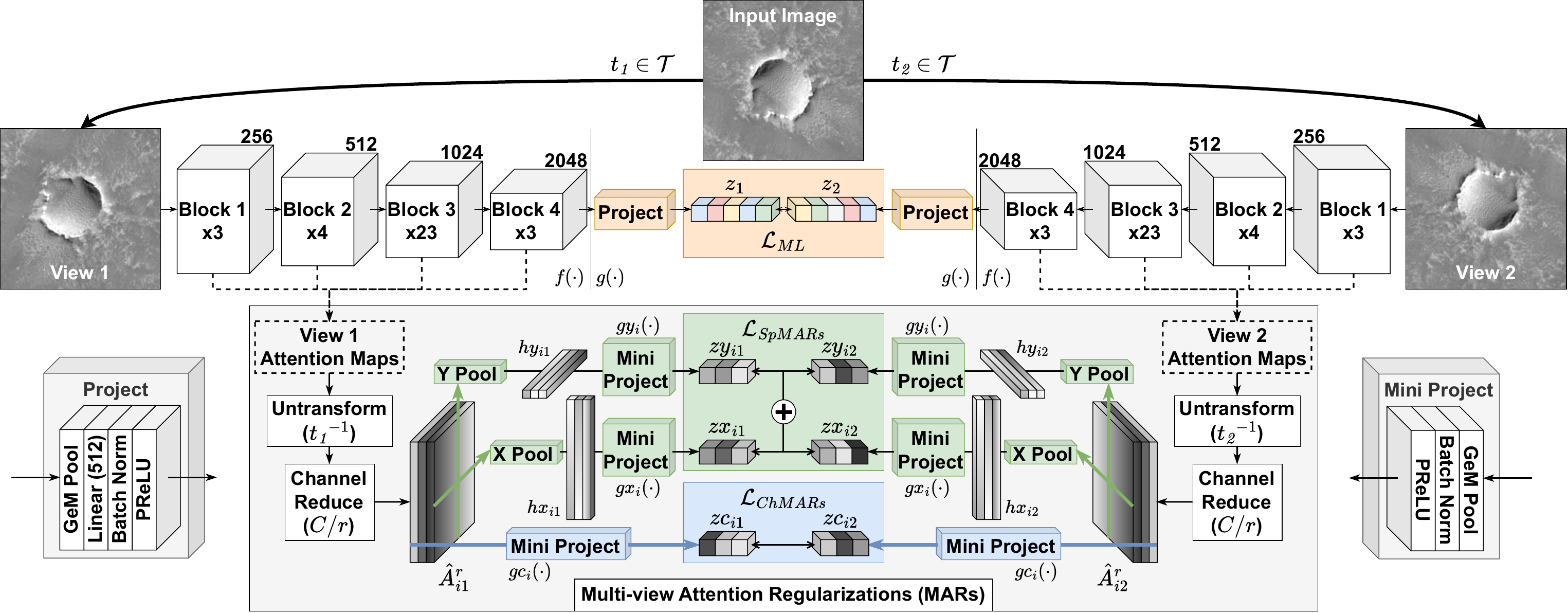}
  \caption{Framework of the proposed Multi-view Attention Regularizations (MARs). MARs aligns the \textit{what} (channel) and \textit{where} (spatial) focus of attention between multiple patch-feature views using distinct metric spaces.}
  \label{fig:arch}
\end{figure*}

\underline{\textbf{Multi-view Attention Regularizations (MARs):}} Once embedded, we regulate the channel and spatial attention focus using a cosine similarity loss:
\begin{equation}\label{eq:cs}
    \mathcal{L}_{\text{cs}}(z_{1},z_{2}) = 1 - \frac{z_{1} \cdot z_{2}}{\lVert z_{1} \rVert_2 \cdot \lVert z_{2} \rVert_2}
\end{equation}
given embeddings $z_1$ and $z_2$. We define a channel-wise MARs ($\mathcal{L}_{\text{ChMARs}}$) as the cosine similarity between channel attention embeddings:
\begin{equation}\label{eq:chmars}
    \mathcal{L}_{\text{ChMARs}}(\hat{A}^r_{i1},\hat{A}^r_{i2}) = \mathcal{L}_{\text{cs}}(zc_{i1},zc_{i2})
\end{equation}
given positive pair, pose-normalized and dimensionality reduced attention maps $\hat{A}^r_{i1}$ and $\hat{A}^r_{i2}$. Likewise, we define a spatial-wise MARs ($\mathcal{L}_{\text{SpMARs}}$) as the cosine similarity between Y-pooled and X-pooled attention embeddings:
\begin{equation}\label{eq:spmars}
    \mathcal{L}_{\text{SpMARs}}(\hat{A}^r_{i1},\hat{A}^r_{i2}) = \mathcal{L}_{\text{cs}}(zy_{i1},zy_{i2}) + \mathcal{L}_{\text{cs}}(zx_{i1},zx_{i2})
\end{equation}
and our combined MARs regularization loss by:
\begin{equation}\label{eq:mars}
    \mathcal{L}_{\text{MARs}}(\hat{A}^r_{i1},\hat{A}^r_{i2}) = \gamma_{Ch} \mathcal{L}_{\text{ChMARs}}(\hat{A}^r_{i1},\hat{A}^r_{i2}) + \gamma_{Sp} \mathcal{L}_{\text{SpMARs}}(\hat{A}^r_{i1},\hat{A}^r_{i2})
\end{equation}
where $\gamma_{Ch}$ and $\gamma_{Sp}$ are weight parameters that control the influence of channel and spatial attention alignment respectively. With augmented image batch $X$ and mined indices $(ap,p,an,n)$ our complete learning objective is given as:
\begin{equation}\label{eq:total}
    \mathcal{L}(X, (ap,p,an,n)) = \mathcal{L}_{\text{ML}}(g(f(X)), (ap,p,an,n)) + \sum_{i=1}^{N \in f(\cdot)} \mathcal{L}_{\text{MARs}}(\hat{A}^r_{ap},\hat{A}^r_{p})
\end{equation}
where
\begin{equation}
    \hat{A}^r_{ap} = \text{Conv}_i^{1}(t_{ap}^{-1}(f_i(X_{ap}))), \quad \hat{A}^r_{p} = \text{Conv}_i^{1}(t_{p}^{-1}(f_i(X_{p})))
\end{equation}
with $X_{ap}$ and $X_p$ the anchor-positive and positive pair images and $f_i(\cdot)$ the $i$'th block in $f(\cdot)$ that outputs attention maps $A_{i}$. Our end-to-end pipeline with MARs regularization is shown in~\autoref{fig:arch}.
\section{Evaluation}\label{sec:eval}
We wish to study lightweight single-shot TRN landmark description using modern metric learning both with and without MARs as well as the effect of different attention and equivariant mechanisms. We first discuss the datasets used for experimentation in this section, followed by a description of our experiments, implementation details, and analysis of the results.

\clearpage

\setlength\intextsep{0pt}
\begin{wrapfigure}{r}{0.25\textwidth}
\centering
\begin{subfigure}{\linewidth}
  \centering
  \includegraphics[width=\linewidth]{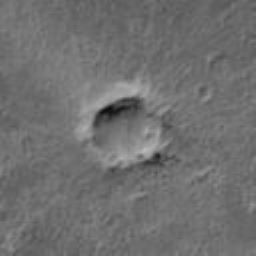}
  \caption{HiRISE~\cite{hirise}}
\end{subfigure}%
\par
\begin{subfigure}{\linewidth}
    \centering
\includegraphics[width=\linewidth]{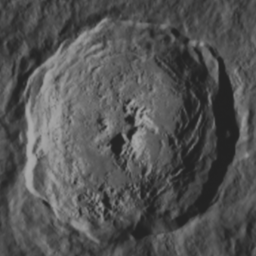}
  \caption{Luna-1}
\end{subfigure}
\caption{Mars (a), Moon (b) landmark examples.}
\label{fig:data}
\end{wrapfigure}

\subsection{Datasets}
We leverage three datasets of Mars, Moon, and Earth landmark images. HiRISE~\cite{hirise} contains 700 Mars crater images and is the only real-world dataset available at the time of writing. We further introduce Luna-1, a 5,067 sample Moon crater dataset generated in the Blender 3D software~\cite{blender} with real-world NASA data products. Luna-1 additionally contains 2,161 emulated navigation frames from a Lunar Reconnaissance Orbiter (LRO) three-orbit reference navigation sequence. An example landmark image from HiRISE and Luna-1 is shown in~\autoref{fig:data}. Additional Luna-1 details and visualizations can be found in the supplementary. For Earth landmarks, we utilize the stadium class from RESISC45~\cite{resisc}, a terrestrial remote sensing scene classification dataset with 700 samples. We refer to HiRISE, Luna-1, and RESISC45 as \textit{Mars Crater}, \textit{Moon Crater}, and \textit{Earth Stadium} respectively. For all datasets, we partition two instance-distinct groups for training and testing such that each group contains half of the available images (as is standard in the literature). In the case of Luna-1, we ensure all craters seen during the navigation sequence are added to the test set before this partitioning.

\subsection{Experiments}
We perform two experiments that emulate landmark recognition behavior during TRN, including a sequential, incremental recall experiment (\textit{Incremental Recall@1}) and an object detection-style description experiment on the navigation frames from Luna-1 (\textit{Moon Navigation}). Additionally, we perform a traditional \textit{Recall@1} (gallery size one) as well as a Luna-1 relocalization experiment (\textit{Moon Lost-in-Space}). Details of these experiments are provided below. Additional results including model execution times on spacecraft hardware and MARs training curves can be viewed in the supplementary.

\underline{\textbf{Incremental Recall@1:}} The embedding database starts empty and test-partition landmarks are randomly selected. Each landmark is augmented by a transform sampled from $\mathcal{T}$. Embeddings are generated single-shot from the model and the database is searched for correspondence. Embeddings are stored in the database if no match is found. We compute Recognition Accuracy (RA) as the percentage of correct matches relative to the total number of landmark matches (either correct, incorrect, or missed). Missed matches are landmarks that were added to the database more than once (i.e., duplicate embeddings). The RA formulation is given in~\autoref{eq:ra}. To provide multiple observations we repeat each landmark in the test partition twice. 
\begin{equation}\label{eq:ra}
    RA = (\frac{\text{Correct Matches}}{\text{Correct Matches} + \text{Incorrect Matches} + \text{Missed Matches}})*100
\end{equation}

\underline{\textbf{Moon Navigation:}} Luna-1 navigation frames are iterated sequentially, where each frame comes paired with ground-truth bounding box annotations of visible craters. For each frame, we first perform non-maximum suppression (NMS) to emulate the use of an object detector (akin YOLO~\cite{yolo}). Landmarks are given by cropping the resulting set of bounding boxes which are in turn augmented by a random transform sampled from $\mathcal{T}$. Embeddings are generated single-shot by the model and RA performance is measured identically to the Incremental Recall@1 experiment. Models trained on Mars Crater are not considered in this experiment due to the domain shift between Mars and Moon. However, one may expect a level of feature generality on crater landmarks from any environment and we report such results in the supplementary.

\underline{\textbf{Moon Lost-in-Space:}} This experiment emulates the kidnapped robot problem in traditional robotics literature. The embedding database is first seeded with all crater landmarks seen during the first orbit of the Luna-1 navigation, where landmarks are detected and augmented identically to the Moon Navigation experiment. Frames from the last orbit are then randomly selected and the RA is reported by matching computed embeddings to those in the database. The database is not updated throughout the experiment outside of the initial seed. Similar to Moon Navigation we only consider models trained on Moon Crater data here, and report a Mars Crater training study in the supplementary. Furthermore, an ablation study over singular transformation types in the family $\mathcal{T}$ for this experiment as well as Moon Navigation is given in the supplementary.

\setlength\intextsep{0pt}
\begin{wraptable}{o}{0.35\textwidth}
\centering
\caption{Evaluated variants of the baseline model. \textit{conv2d:} PyTorch convolution. \textit{RIC:} Rotation Invariant Convolution~\cite{ric}. \textit{SE:} Squeeze-Excitation attention~\cite{se}. \textit{CA:} Coordinate Attention~\cite{ca}.}
\resizebox{\linewidth}{!}{%
\begin{tabular}{@{}c|ccc@{}}
\toprule
Name & Conv. & Att. & Loss     \\ \midrule
conv2d SE  & conv2d      & SE        & $\ml$ Only \\
RIC CA     & RIC       & CA        & $\ml$ Only \\
MARs       & RIC       & CA        & $\mars$    \\ \bottomrule
\end{tabular}%
}
\label{tab:models}
\end{wraptable}

\subsection{Implementation Details}
To determine the effectiveness of metric learning for robust landmark description, it is imperative to understand two primary conditions for TRN: (i) recognition over time with many similar terrain features encountered sequentially, and (ii) the unique transformation space in remote sensing. Therefore, we frame this study as a measure of modern metric learning invariancy and discriminative properties under these conditions and elect not to compare against fully-fledged Earth-based systems that are unsuitable for onboard spaceflight. Additionally, we seek to understand the influence of MARs on various metric learning losses and the effect of different attention and equivariant setups. 

We evaluate three variants of the baseline model which are described in~\autoref{tab:models}. Each model effectively adds a level of equivariance (and in theory, robustness to challenging multi-view appearance change) from the last; i.e., \textit{conv2d SE} learned equivariance only, \textit{RIC CA} learned and explicit equivariance, \textit{MARs} learned and explicit equivariance with attention constraints. We study the effects of each model across nine discriminative learning losses ($\ml$) found recent in the literature, including Circle Loss~\cite{circle}, Direction-Regularized Multi-Similarity (DR-MS)~\cite{deen}, NTXent~\cite{ntxent}, PNP~\cite{pnp}, Proxy Anchor~\cite{proxyanchor}, ProxyNCA++~\cite{proxynca++}, Subcenter ArcFace~\cite{subcenter}, Supervised Contrastive (SupCon)~\cite{supcon}, and Proxy Synthesis~\cite{synproxy}. We train all models for 150 epochs using a batch size of 32 on Earth Stadium and Mars Crater and 128 on Moon Crater. 

We use the PyTorch Metric Learning (PML)~\cite{pytorchml} library for MS miner implementation as well as all $\ml$ losses except ProxyNCA++ and Proxy Synthesis, in which we use the paper implementations. The Faiss~\cite{faiss} library is used as the embedding database in all experiments. We use cosine similarity for database comparison and define a matching threshold of 0.9. If multiple embeddings are retrieved above this threshold we consider the largest one a match. In Moon Navigation an NMS threshold of 0.5 is used based on the YOLO default. The $p$ parameter in GeM layers is learnable with an initial value of $3$. The reduction factor $r$ is set to $4$. All $\mars$ models have $\gamma_{Ch}$ and $\gamma_{Sp}$ parameters set to 0.15.

\subsection{Experimental Results}
\begin{table*}
\centering
\caption{Recall@1 (gallery size one) and Incremental Recall@1 recognition accuracy for all models and $\ml$ losses. Bold values signify the highest performing model for each $\ml$, while underlined values show the best model/$\ml$ variant on each dataset.}
\resizebox{\textwidth}{!}{%
\begin{tabular}{@{}l|ccccccccc|ccccccccc@{}}
\toprule
\multicolumn{1}{c|}{} &
  \multicolumn{9}{c|}{\textbf{Recall@1}} &
  \multicolumn{9}{c}{\textbf{Incremental Recall@1}} \\ \cmidrule(l){2-19} 
\multicolumn{1}{c|}{} &
  \multicolumn{3}{c|}{Earth Stadium} &
  \multicolumn{3}{c|}{Mars Crater} &
  \multicolumn{3}{c|}{Moon Crater} &
  \multicolumn{3}{c|}{Earth Stadium} &
  \multicolumn{3}{c|}{Mars Crater} &
  \multicolumn{3}{c}{Moon Crater} \\ \cmidrule(l){2-19} 
\multicolumn{1}{c|}{\multirow{-4}{*}{$\mathcal{L}_{\text{ML}}$}} &
  \begin{tabular}[c]{@{}c@{}}conv2d\\SE\end{tabular} &
  \begin{tabular}[c]{@{}c@{}}RIC\\CA\end{tabular} &
  \multicolumn{1}{c|}{\begin{tabular}[c]{@{}c@{}}MARs\\(Ours)\end{tabular}} &
  \begin{tabular}[c]{@{}c@{}}conv2d\\SE\end{tabular} &
  \begin{tabular}[c]{@{}c@{}}RIC\\CA\end{tabular} &
  \multicolumn{1}{c|}{\begin{tabular}[c]{@{}c@{}}MARs\\(Ours)\end{tabular}} &
  \begin{tabular}[c]{@{}c@{}}conv2d\\SE\end{tabular} &
  \begin{tabular}[c]{@{}c@{}}RIC\\CA\end{tabular} &
  \begin{tabular}[c]{@{}c@{}}MARs\\(Ours)\end{tabular} &
  \begin{tabular}[c]{@{}c@{}}conv2d\\SE\end{tabular} &
  \begin{tabular}[c]{@{}c@{}}RIC\\CA\end{tabular} &
  \multicolumn{1}{c|}{\begin{tabular}[c]{@{}c@{}}MARs\\(Ours)\end{tabular}} &
  \begin{tabular}[c]{@{}c@{}}conv2d\\SE\end{tabular} &
  \begin{tabular}[c]{@{}c@{}}RIC\\CA\end{tabular} &
  \multicolumn{1}{c|}{\begin{tabular}[c]{@{}c@{}}MARs\\(Ours)\end{tabular}} &
  \begin{tabular}[c]{@{}c@{}}conv2d\\SE\end{tabular} &
  \begin{tabular}[c]{@{}c@{}}RIC\\CA\end{tabular} &
  \begin{tabular}[c]{@{}c@{}}MARs\\(Ours)\end{tabular} \\ \midrule
\tcircle &
  86.29 &
  \textbf{93.71} &
  \multicolumn{1}{c|}{90.29} &
  59.71 &
  \textbf{92.29} &
  \multicolumn{1}{c|}{80.57} &
  \textbf{98.38} &
  97.75 &
  96.93 &
  4.79 &
  5.11 &
  \multicolumn{1}{c|}{\textbf{5.13}} &
  3.59 &
  \textbf{60.81} &
  \multicolumn{1}{c|}{12.56} &
  43.36 &
  \textbf{60.09} &
  30.46 \\
\tdrms &
  86.57 &
  88.86 &
  \multicolumn{1}{c|}{\textbf{89.71}} &
  66.57 &
  \textbf{84.29} &
  \multicolumn{1}{c|}{62.57} &
  \textbf{96.14} &
  85.97 &
  96.02 &
  4.33 &
  5.27 &
  \multicolumn{1}{c|}{\textbf{54.32}} &
  4.04 &
  \textbf{48.48} &
  \multicolumn{1}{c|}{4.04} &
  62.55 &
  2.47 &
  \textbf{64.57} \\
\tntxent &
  \textbf{95.43} &
  91.71 &
  \multicolumn{1}{c|}{94.29} &
  66.57 &
  83.43 &
  \multicolumn{1}{c|}{\textbf{91.71}} &
  98.03 &
  98.98 &
  \textbf{99.57} &
  \textbf{12.23} &
  7.75 &
  \multicolumn{1}{c|}{8.59} &
  4.49 &
  \textbf{34.10} &
  \multicolumn{1}{c|}{34.01} &
  69.85 &
  77.06 &
  \textbf{81.69} \\
\tpnp &
  80.57 &
  79.43 &
  \multicolumn{1}{c|}{\textbf{85.14}} &
  43.71 &
  \textbf{80.86} &
  \multicolumn{1}{c|}{68.00} &
  75.18 &
  88.65 &
  \textbf{94.09} &
  4.64 &
  \textbf{5.27} &
  \multicolumn{1}{c|}{4.66} &
  3.14 &
  \textbf{16.70} &
  \multicolumn{1}{c|}{5.14} &
  10.24 &
  15.54 &
  \textbf{40.84} \\
\tproxyanchor &
  90.57 &
  99.71 &
  \multicolumn{1}{c|}{\underline{\textbf{100.00}}} &
  56.86 &
  84.86 &
  \multicolumn{1}{c|}{\textbf{98.57}} &
  99.96 &
  -- &
  \underline{\textbf{100.00}} &
  4.48 &
  72.86 &
  \multicolumn{1}{c|}{\underline{\textbf{78.06}}} &
  3.44 &
  10.45 &
  \multicolumn{1}{c|}{\underline{\textbf{71.10}}} &
  94.56 &
  -- &
  \underline{\textbf{94.78}} \\
\tproxynca &
  95.71 &
  96.29 &
  \multicolumn{1}{c|}{\textbf{99.71}} &
  63.14 &
  \underline{\textbf{99.43}} &
  \multicolumn{1}{c|}{78.57} &
  99.65 &
  \textbf{98.78} &
  95.47 &
  4.33 &
  4.64 &
  \multicolumn{1}{c|}{\textbf{12.00}} &
  4.04 &
  \textbf{7.71} &
  \multicolumn{1}{c|}{4.79} &
  56.03 &
  \textbf{71.23} &
  70.27 \\
\tsubcenter &
  77.71 &
  \textbf{87.71} &
  \multicolumn{1}{c|}{83.43} &
  40.86 &
  62.29 &
  \multicolumn{1}{c|}{\textbf{79.14}} &
  -- &
  -- &
  \textbf{94.05} &
  \textbf{4.49} &
  4.48 &
  \multicolumn{1}{c|}{4.19} &
  3.29 &
  4.99 &
  \multicolumn{1}{c|}{\textbf{38.69}} &
  -- &
  -- &
  \textbf{20.45} \\
\tsupcon &
  76.57 &
  92.00 &
  \multicolumn{1}{c|}{\textbf{95.71}} &
  74.29 &
  \textbf{94.86} &
  \multicolumn{1}{c|}{91.43} &
  97.08 &
  \textbf{99.21} &
  98.98 &
  4.19 &
  5.80 &
  \multicolumn{1}{c|}{\textbf{46.43}} &
  4.65 &
  \textbf{57.52} &
  \multicolumn{1}{c|}{49.19} &
  16.73 &
  79.42 &
  \textbf{84.11} \\
\tsynproxy &
  94.86 &
  95.14 &
  \multicolumn{1}{c|}{\textbf{99.71}} &
  76.86 &
  74.57 &
  \multicolumn{1}{c|}{\textbf{99.14}} &
  99.68 &
  \textbf{99.96} &
  99.84 &
  4.33 &
  \textbf{39.41} &
  \multicolumn{1}{c|}{22.57} &
  4.03 &
  4.97 &
  \multicolumn{1}{c|}{\textbf{35.14}} &
  \textbf{91.40} &
  64.71 &
  17.47 \\ \bottomrule
\end{tabular}%
}
\label{tab:recall_testrec}
\end{table*}
\underline{\textbf{Recall@1:}}~\autoref{tab:recall_testrec} (left) displays results for Recall@1. Firstly, including explicit rotation equivariance and spatial attention (RIC CA) leads to improvements on many $\ml$, suggesting that learning transformation robustness alone is not enough and a combination of learned and explicit equivariance is necessary. On Mars Crater data, MARs leads to substantial improvements on certain $\ml$ such as NTXent, Proxy Anchor, Subcenter ArcFace, and Proxy Synthesis which were improved by roughly 10\%, 16\%, 27\%, and 33\% respectively compared to RIC CA. This is evidence that attention similarity heavily influences $f(\cdot)$ feature selection and results in more discriminative embeddings. Conv2d SE and RIC CA variants for Subcenter ArcFace on Moon Crater see a failure to converge during training while MARs variants do not. This is significant as it demonstrates a boost in representation power that can enable $\ml$ losses that would otherwise fail. Overall, a MARs model variant is best-in-class for Earth Stadium and Moon Carter, and competitive (<1\% difference) on Mars Crater.

Benefits are not guaranteed as we observe lower accuracy than RIC CA with MARs for certain $\ml$ such as DR-MS on Mars Crater, which sees a performance decrease of roughly 26\%. This reveals a correlation between attention alignment and embedding separability that is $\ml$ specific. Knowing how attention information ultimately presents itself in the embedding projected by $g(\cdot)$ is not obvious, which may indicate incompatible $\ml$ where attention information is ultimately obscured and not readily distinguishable in the $\ml$ space.

\underline{\textbf{Incremental Recall@1:}} Recognition accuracy for the Incremental Recall@1 experiment is shown in~\autoref{tab:recall_testrec} (right). We see very low accuracy with conv2d SE on Earth Stadium and Mars Crater, which signals poor representation power and an indiscernible metric space on these smaller datasets. Encoding rotational equivariance in RIC CA helps alleviate this issue in cases such as Proxy Anchor on Earth Stadium and many $\ml$ on Mars Crater. MARs has a profound impact in cases where RIC CA offers little to no benefit, such as Proxy Anchor on Mars Crater where we see a roughly 85\% improvement over RIC CA. On Moon Crater, MARs attention constraints improve 6/9 $\ml$ losses and is competitive on Subcenter ArcFace ($<1\%$ difference), indicating benefits with more training data. Furthermore, we see a similar pattern of behavior to the Recall@1 experiment where the performance of RIC CA and MARs varies wildly across $\ml$ losses, as shown by Proxy Synthesis on Mars/Moon Crater. Overall, Proxy Anchor with MARs is best-in-class on every dataset for this experiment.

\underline{\textbf{Moon Navigation and Lost-in-Space:}}~\autoref{tab:lro} displays results for Moon Navigation (left) and Moon Lost-in-Space (right). For Moon Navigation, conv2d SE retains maximum performance on 4/9 $\ml$ losses while RIC CA achieves the highest accuracy only on SupCon loss, supporting the theory that fully explicit\begin{wraptable}[19]{o}{0.5\textwidth}
\centering
\caption{Accuracy for Moon Navigation (left) and Moon Lost-in-Space (right).}
\resizebox{\linewidth}{!}{%
\begin{tabular}{@{}l|ccc|ccc@{}}
\toprule
\multicolumn{1}{c|}{\multirow{3}{*}{$\mathcal{L}_{\text{ML}}$}} &
  \multicolumn{3}{c|}{\begin{tabular}[c]{@{}c@{}}\textbf{Moon Navigation}\end{tabular}} &
  \multicolumn{3}{c}{\begin{tabular}[c]{@{}c@{}}\textbf{Moon Lost-in-Space}\end{tabular}} \\ \cmidrule(l){2-7} 
\multicolumn{1}{c|}{} &
  \begin{tabular}[c]{@{}c@{}}conv2d\\SE\end{tabular} &
  \begin{tabular}[c]{@{}c@{}}RIC\\CA\end{tabular} &
  \begin{tabular}[c]{@{}c@{}}MARs\\(Ours)\end{tabular} &
  \begin{tabular}[c]{@{}c@{}}conv2d\\SE\end{tabular} &
  \begin{tabular}[c]{@{}c@{}}RIC\\CA\end{tabular} &
  \begin{tabular}[c]{@{}c@{}}MARs\\(Ours)\end{tabular} \\ \midrule
\tcircle      & \textbf{58.07} & 37.97          & 38.46          & 94.03          & \textbf{96.68} & 92.31          \\
\tdrms        & \textbf{37.69} & 3.12           & 36.68          & 86.34          & 88.06          & \textbf{90.05} \\
\tntxent      & 48.25          & 32.00          & \textbf{57.68} & 94.83          & 83.16          & \textbf{96.29} \\
\tpnp         & 14.34          & 23.34          & \textbf{24.66} & 61.41          & \textbf{77.98} & 75.46          \\
\tproxyanchor & 64.17          & --             & \underline{\textbf{66.31}} & \underline{\textbf{97.21}} & --             & 96.02          \\
\tproxynca    & \textbf{58.27} & 53.92          & 35.87          & \textbf{94.69} & 93.24          & 91.38          \\
\tsubcenter   & --             & --             & \textbf{40.63} & --             & --             & \textbf{81.17} \\
\tsupcon      & 17.92          & \textbf{42.28} & 37.50          & 89.39          & 90.32          & \textbf{90.58} \\
\tsynproxy    & \textbf{61.26} & 60.53          & 32.67          & \textbf{96.42} & 93.77          & 36.87          \\ \bottomrule
\end{tabular}%
}
\label{tab:lro}

\caption{MARs $\gamma$ ablation study.}
\resizebox{0.5\textwidth}{!}{%
\begin{tabular}{@{}cc|cccc@{}}
\toprule
$\gamma_{Ch}$ &
$\gamma_{Sp}$ &
  Recall@1 &
  \begin{tabular}[c]{@{}c@{}}Incremental\\Recall@1\end{tabular} &
  \begin{tabular}[c]{@{}c@{}}Moon\\Navigation\end{tabular} &
  \begin{tabular}[c]{@{}c@{}}Moon\\Lost-in-Space\end{tabular} \\ \midrule
0.0  & 0.3  & \textbf{100}    & 59.429 & \textbf{48.204} & 89.594 \\
0.15 & 0.15 & 98.571 & \textbf{71.105} & 38.301 & 89.335 \\
0.3  & 0.0  & 69.43  & 4.18   & 1.93   & 62.62  \\
1    & 1    & 96.571 & 61.517 & 23.593 & \textbf{89.733} \\ \bottomrule
\end{tabular}%
}
\label{tab:ablation}
\end{wraptable}equivariance may be undesirable in cases with more training data, and a partial (i.e., learned) equivariance is better suited due to the sufficiency of the learning framework to distinguish low-level features under transformation~\cite{leq3}. Nevertheless, MARs is quite competitive in instances where RIC CA has worse performance than conv2d SE showing that the additional \textit{learned} multi-view attention constraint counteracts the negative effects of the explicit properties. In total, MARs achieves the highest performance on 4/9 $\ml$ and obtains best-in-class with Proxy Anchor. Accuracy is relatively high for all methods on Moon Lost-in-Space, which could be an artifact of the high-framerate navigation sequence that fills the embedding database with many duplicate (although augmented) craters throughout the first orbit. MARs demonstrates performance increases on 4/9 $\ml$ losses once again with only NTXent and Subcenter ArcFace being common among both experiments. 

\underline{\textbf{Ablation Study, $\gamma$ Parameter:}}~\autoref{tab:ablation} gives an ablation study over the $\gamma$ parameters in MARs. We train four combinations of $\gamma_{Ch}$ and $\gamma_{Sp}$ using Proxy Anchor $\ml$ on Mars Crater, where we can see a clear sensitivity. Spatial attention has the biggest impact where $\gamma_{Sp} = 0$ reduces accuracy in all experiments, indicating the importance of spatial attention alignment in multi-view learning. The conservative 0.15 for both $\gamma_{Ch}$ and $\gamma_{Sp}$ gives utility to the channel component, as we see the best results on Recall@1 and Incremental Recall@1 accuracy (and only slightly less accuracy on Moon Lost-in-Space). A perhaps surprising result is the difference in performance (or lack thereof) between low parameter settings (0.15) and unweighted settings ($\gamma_{Ch} = \gamma_{Sp} = 1$), implying that the magnitude of $\mars$ has little effect on the optimization.

\begin{figure*}
\centering
\begin{subfigure}{.5\textwidth}
  \centering
  \includegraphics[width=0.98\linewidth]{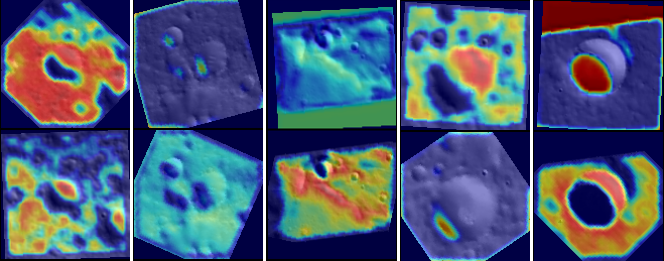}
  \caption{RIC CA}
\end{subfigure}%
\begin{subfigure}{.5\textwidth}
    \centering
\includegraphics[width=0.98\linewidth]{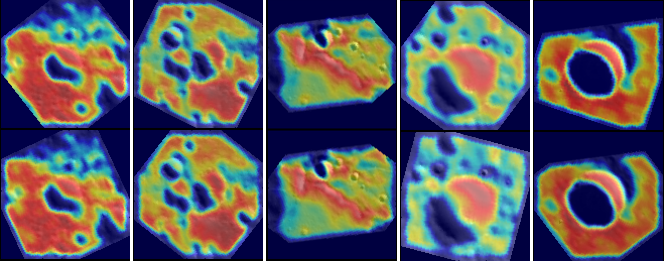}
  \caption{MARs}
\end{subfigure}
\caption{Attention visualizations with EigenCAM~\cite{eigencam} on Mars Crater trained with Proxy Anchor $\ml$.}
\label{fig:grid}
\end{figure*}

\underline{\textbf{Qualitative Analysis:}} 
\autoref{fig:grid} visualizes attention focus via the EigenCAM~\cite{eigencam} algorithm, comparing RIC CA and MARs ($\gamma_{Sp} = 0.3, \gamma_{Ch} = 0$) for Proxy Anchor on Mars Crater. MARs shows a near-identical magnitude EigenCAM between each positive pair of pose-normalized landmarks where RIC CA focuses on disparate regions. Additional examples and animations of attention evolution during the training can be viewed in the supplementary.

\setlength\intextsep{12pt}

\section{Conclusion}\label{sec:conc}
The utility of metric learning as a single-shot landmark description technique for spacecraft TRN was thoroughly explored in this work. We demonstrated that metric learning alone cannot adequately descript fine-grained instances of celestial terrain given multi-view observations and complex transformation spaces. We show that traditional workarounds such as equivariant convolutional layers are in many cases still insufficient. We identify shortcomings with the view-unaware attention mechanism and proposed Multi-view Attention Regularizations (MARs) to regulate attention focus between views. MARs enacts a soft learning constraint that prevents attention collapse, effectively driving the \textit{what} and \textit{where} elements of attention together and eases the downstream separability task. We demonstrated the utility of our method through rigorous and comprehensive experimentation, where we showed regular improvements to a wide range of metric learning losses by upwards of 85\% on navigation-style tasks. We additionally introduced the Luna-1 dataset to facilitate more active research in TRN landmark recognition, consisting of photo-realistic Moon crater landmarks and paired navigation images using real-world NASA data.

% ---- Bibliography ----
%
% BibTeX users should specify bibliography style 'splncs04'.
% References will then be sorted and formatted in the correct style.
%
\bibliographystyle{splncs04}
\bibliography{refs}
\end{document}

% ---------------------------------------------------------------
% \title{MARs: Multi-view Attention Regularizations for Space Terrain Recognition}
\title{Supplementary Material for MARs: Multi-view Attention Regularizations for Patch-based Feature Recognition of Space Terrain}

% TODO REVIEW: If the paper title is too long for the running head, you can set
% an abbreviated paper title here. If not, comment out.
\titlerunning{MARs: Multi-view Attention Regularizations}

% TODO FINAL: Replace with your author list. 
% Include the authors' OCRID for the camera-ready version, if at all possible.
\author{Timothy Chase Jr\orcidlink{0009-0004-3075-0790} \and
Karthik Dantu\orcidlink{0000-0002-7497-6722}}

% TODO FINAL: Replace with an abbreviated list of authors.
\authorrunning{T.~Chase Jr, K.~Dantu}
% First names are abbreviated in the running head.
% If there are more than two authors, 'et al.' is used.

% TODO FINAL: Replace with your institution list.
\institute{University at Buffalo \\
\email{\{tbchase,kdantu\}@buffalo.edu}}

\maketitle

% Please add the following required packages to your document preamble:
% \usepackage{booktabs}
% \usepackage{multirow}
% \usepackage{graphicx}
\begin{table*}[p]
\centering
\caption{Moon Navigation accuracy for models trained on Moon Crater, ablated by transformation type.}
\resizebox{0.825\textwidth}{!}{%
\begin{tabular}{@{}l|ccc|ccc|ccc|ccc@{}}
\toprule
\multicolumn{1}{c|}{\multirow{3}{*}{$\mathcal{L}_{\text{ML}}$}} &
  \multicolumn{3}{c|}{Illumination} &
  \multicolumn{3}{c|}{Translation} &
  \multicolumn{3}{c|}{Rotation} &
  \multicolumn{3}{c}{All} \\ \cmidrule(l){2-13} 
\multicolumn{1}{c|}{} &
  \begin{tabular}[c]{@{}c@{}}conv2d\\SE\end{tabular} &
  \begin{tabular}[c]{@{}c@{}}RIC\\CA\end{tabular} &
  \begin{tabular}[c]{@{}c@{}}MARs\\(Ours)\end{tabular} &
  \begin{tabular}[c]{@{}c@{}}conv2d\\SE\end{tabular} &
  \begin{tabular}[c]{@{}c@{}}RIC\\CA\end{tabular} &
  \begin{tabular}[c]{@{}c@{}}MARs\\(Ours)\end{tabular} &
  \begin{tabular}[c]{@{}c@{}}conv2d\\SE\end{tabular} &
  \begin{tabular}[c]{@{}c@{}}RIC\\CA\end{tabular} &
  \begin{tabular}[c]{@{}c@{}}MARs\\(Ours)\end{tabular} &
  \begin{tabular}[c]{@{}c@{}}conv2d\\SE\end{tabular} &
  \begin{tabular}[c]{@{}c@{}}RIC\\CA\end{tabular} &
  \begin{tabular}[c]{@{}c@{}}MARs\\(Ours)\end{tabular} \\ \midrule
\tcircle &
  \textbf{85.20} &
  37.47 &
  56.23 &
  \underline{\textbf{76.28}} &
  39.81 &
  50.83 &
  \textbf{75.91} &
  41.57 &
  49.92 &
  \textbf{58.07} &
  37.97 &
  38.46 \\
\tdrms &
  \textbf{44.26} &
  3.82 &
  44.25 &
  41.70 &
  3.93 &
  \textbf{42.27} &
  \textbf{43.32} &
  3.91 &
  42.89 &
  \textbf{37.69} &
  3.12 &
  36.68 \\
\tntxent &
  53.67 &
  40.95 &
  \textbf{69.33} &
  53.50 &
  36.20 &
  \textbf{64.25} &
  54.35 &
  39.24 &
  \textbf{67.00} &
  48.25 &
  32.00 &
  \textbf{57.68} \\
\tpnp &
  14.44 &
  25.03 &
  \textbf{29.27} &
  14.40 &
  23.27 &
  \textbf{26.46} &
  16.30 &
  28.95 &
  \textbf{31.31} &
  14.34 &
  23.34 &
  \textbf{24.66} \\
\tproxyanchor &
  72.23 &
  -- &
  \textbf{83.02} &
  66.83 &
  -- &
  \textbf{72.02} &
  71.62 &
  -- &
  \textbf{79.20} &
  64.17 &
  -- &
  \underline{\textbf{66.31}} \\
\tproxynca &
  \textbf{80.82} &
  67.48 &
  36.01 &
  \textbf{75.98} &
  62.48 &
  34.17 &
  \textbf{77.45} &
  69.69 &
  40.14 &
  \textbf{58.27} &
  53.92 &
  35.87 \\
\tsubcenter &
  -- &
  -- &
  \textbf{61.59} &
  -- &
  -- &
  \textbf{60.54} &
  -- &
  -- &
  \textbf{64.02} &
  -- &
  -- &
  \textbf{40.63} \\
\tsupcon &
  17.24 &
  \textbf{54.03} &
  43.78 &
  17.45 &
  \textbf{46.40} &
  39.16 &
  17.25 &
  \textbf{52.65} &
  44.51 &
  17.92 &
  \textbf{42.28} &
  37.50 \\
\tsynproxy &
  69.86 &
  \underline{\textbf{86.38}} &
  83.94 &
  65.07 &
  \textbf{75.56} &
  67.24 &
  68.89 &
  \underline{\textbf{82.98}} &
  72.99 &
  \textbf{61.26} &
  60.53 &
  32.67 \\ \bottomrule
\end{tabular}%
}
\label{tab:lrorec_lunar}
\end{table*} 
% Please add the following required packages to your document preamble:
% \usepackage{booktabs}
% \usepackage{multirow}
% \usepackage{graphicx}
\begin{table*}[p]
\centering
\caption{Moon Navigation accuracy for models trained on Mars Crater, ablated by transformation type.}
\resizebox{0.825\textwidth}{!}{%
\begin{tabular}{@{}l|ccc|ccc|ccc|ccc@{}}
\toprule
\multicolumn{1}{c|}{\multirow{3}{*}{$\mathcal{L}_{\text{ML}}$}} &
  \multicolumn{3}{c|}{Illumination} &
  \multicolumn{3}{c|}{Translation} &
  \multicolumn{3}{c|}{Rotation} &
  \multicolumn{3}{c}{All} \\ \cmidrule(l){2-13} 
\multicolumn{1}{c|}{} &
  \begin{tabular}[c]{@{}c@{}}conv2d\\SE\end{tabular} &
  \begin{tabular}[c]{@{}c@{}}RIC\\CA\end{tabular} &
  \begin{tabular}[c]{@{}c@{}}MARs\\(Ours)\end{tabular} &
  \begin{tabular}[c]{@{}c@{}}conv2d\\SE\end{tabular} &
  \begin{tabular}[c]{@{}c@{}}RIC\\CA\end{tabular} &
  \begin{tabular}[c]{@{}c@{}}MARs\\(Ours)\end{tabular} &
  \begin{tabular}[c]{@{}c@{}}conv2d\\SE\end{tabular} &
  \begin{tabular}[c]{@{}c@{}}RIC\\CA\end{tabular} &
  \begin{tabular}[c]{@{}c@{}}MARs\\(Ours)\end{tabular} &
  \begin{tabular}[c]{@{}c@{}}conv2d\\SE\end{tabular} &
  \begin{tabular}[c]{@{}c@{}}RIC\\CA\end{tabular} &
  \begin{tabular}[c]{@{}c@{}}MARs\\(Ours)\end{tabular} \\ \midrule
\tcircle &
  2.13 &
  26.32 &
  \textbf{63.46} &
  3.31 &
  25.85 &
  \textbf{60.32} &
  3.29 &
  28.43 &
  \textbf{69.84} &
  1.75 &
  19.99 &
  \underline{\textbf{47.62}} \\
\tdrms &
  2.40 &
  \textbf{18.04} &
  2.48 &
  2.81 &
  \textbf{13.51} &
  3.30 &
  2.97 &
  \textbf{18.23} &
  3.28 &
  1.98 &
  \textbf{12.25} &
  2.07 \\
\tntxent &
  2.06 &
  \textbf{15.65} &
  14.03 &
  3.45 &
  \textbf{14.43} &
  12.38 &
  3.54 &
  \textbf{16.36} &
  13.96 &
  1.72 &
  \textbf{11.34} &
  9.65 \\
\tpnp &
  1.28 &
  \textbf{6.11} &
  3.70 &
  3.44 &
  \textbf{6.99} &
  3.70 &
  3.55 &
  \textbf{7.72} &
  3.76 &
  1.30 &
  \textbf{5.36} &
  3.05 \\
\tproxyanchor &
  1.88 &
  6.31 &
  \textbf{51.32} &
  3.41 &
  6.54 &
  \textbf{49.96} &
  3.20 &
  7.01 &
  \textbf{57.90} &
  1.48 &
  5.15 &
  \textbf{38.30} \\
\tproxynca &
  1.58 &
  \underline{\textbf{78.94}} &
  2.20 &
  3.26 &
  \underline{\textbf{77.30}} &
  3.57 &
  3.32 &
  \underline{\textbf{78.23}} &
  3.67 &
  1.06 &
  \textbf{40.98} &
  1.62 \\
\tsubcenter &
  1.42 &
  3.36 &
  \textbf{9.08} &
  3.19 &
  5.43 &
  \textbf{9.44} &
  3.24 &
  5.11 &
  \textbf{10.65} &
  1.50 &
  3.12 &
  \textbf{8.03} \\
\tsupcon &
  3.14 &
  \textbf{19.50} &
  17.46 &
  3.40 &
  \textbf{17.05} &
  13.81 &
  3.69 &
  \textbf{19.01} &
  17.35 &
  2.44 &
  \textbf{15.69} &
  10.75 \\
\tsynproxy &
  2.35 &
  4.54 &
  \textbf{61.86} &
  3.54 &
  5.37 &
  \textbf{59.69} &
  3.11 &
  5.42 &
  \textbf{69.17} &
  1.81 &
  3.45 &
  \textbf{47.33} \\ \bottomrule
\end{tabular}%
}
\label{tab:lrorec_hirise}
\end{table*}
% Please add the following required packages to your document preamble:
% \usepackage{booktabs}
% \usepackage{multirow}
% \usepackage{graphicx}
\begin{table*}[p]
\centering
\caption{Moon Lost-in-Space accuracy for models trained on Moon Crater, ablated by transformation type.}
\resizebox{0.825\textwidth}{!}{%
\begin{tabular}{@{}l|ccccccllcllc@{}}
\toprule
\multicolumn{1}{c|}{\multirow{3}{*}{$\mathcal{L}_{\text{ML}}$}} &
  \multicolumn{3}{c}{Illumination} &
  \multicolumn{3}{c}{Translation} &
  \multicolumn{3}{c}{Rotation} &
  \multicolumn{3}{c}{All} \\ \cmidrule(l){2-13} 
\multicolumn{1}{c|}{} &
  \begin{tabular}[c]{@{}c@{}}conv2d\\SE\end{tabular} &
  \begin{tabular}[c]{@{}c@{}}RIC\\CA\end{tabular} &
  \multicolumn{1}{c|}{\begin{tabular}[c]{@{}c@{}}MARs\\(Ours)\end{tabular}} &
  \begin{tabular}[c]{@{}c@{}}conv2d\\SE\end{tabular} &
  \begin{tabular}[c]{@{}c@{}}RIC\\CA\end{tabular} &
  \multicolumn{1}{c|}{\begin{tabular}[c]{@{}c@{}}MARs\\(Ours)\end{tabular}} &
  \multicolumn{1}{c}{\begin{tabular}[c]{@{}c@{}}conv2d\\SE\end{tabular}} &
  \multicolumn{1}{c}{\begin{tabular}[c]{@{}c@{}}RIC\\CA\end{tabular}} &
  \multicolumn{1}{c|}{\begin{tabular}[c]{@{}c@{}}MARs\\(Ours)\end{tabular}} &
  \multicolumn{1}{c}{\begin{tabular}[c]{@{}c@{}}conv2d\\SE\end{tabular}} &
  \multicolumn{1}{c}{\begin{tabular}[c]{@{}c@{}}RIC\\CA\end{tabular}} &
  \begin{tabular}[c]{@{}c@{}}MARs\\ (Ours)\end{tabular} \\ \midrule
\tcircle &
  \textbf{98.81} &
  98.54 &
  \multicolumn{1}{c|}{\textbf{98.81}} &
  \textbf{98.54} &
  97.88 &
  \multicolumn{1}{c|}{98.28} &
  \textbf{97.48} &
  97.35 &
  \multicolumn{1}{c|}{96.55} &
  94.03 &
  \textbf{96.68} &
  92.31 \\
\tdrms &
  \textbf{98.14} &
  \textbf{98.14} &
  \multicolumn{1}{c|}{97.88} &
  \textbf{98.67} &
  97.21 &
  \multicolumn{1}{c|}{97.35} &
  96.42 &
  \textbf{97.08} &
  \multicolumn{1}{c|}{96.82} &
  86.34 &
  88.06 &
  \textbf{90.05} \\
\tntxent &
  98.67 &
  97.61 &
  \multicolumn{1}{c|}{\textbf{98.81}} &
  98.28 &
  96.29 &
  \multicolumn{1}{c|}{\textbf{98.41}} &
  \textbf{97.35} &
  96.15 &
  \multicolumn{1}{c|}{97.21} &
  94.83 &
  83.16 &
  \textbf{96.29} \\
\tpnp &
  96.42 &
  \textbf{97.08} &
  \multicolumn{1}{c|}{\textbf{97.08}} &
  94.69 &
  96.15 &
  \multicolumn{1}{c|}{\textbf{96.68}} &
  95.36 &
  \textbf{95.49} &
  \multicolumn{1}{c|}{95.36} &
  61.41 &
  \textbf{77.98} &
  75.46 \\
\tproxyanchor &
  \underline{\textbf{98.94}} &
  -- &
  \multicolumn{1}{c|}{98.67} &
  \textbf{98.54} &
  -- &
  \multicolumn{1}{c|}{98.01} &
  \textbf{97.75} &
  \multicolumn{1}{c}{--} &
  \multicolumn{1}{c|}{97.61} &
  \underline{\textbf{97.21}} &
  \multicolumn{1}{c}{--} &
  96.02 \\
\tproxynca &
  98.28 &
  98.28 &
  \multicolumn{1}{c|}{\textbf{98.41}} &
  \underline{\textbf{99.07}} &
  98.41 &
  \multicolumn{1}{c|}{98.14} &
  \underline{\textbf{98.14}} &
  97.61 &
  \multicolumn{1}{c|}{97.75} &
  \textbf{94.69} &
  93.24 &
  91.38 \\
\tsubcenter &
  -- &
  -- &
  \multicolumn{1}{c|}{\textbf{98.01}} &
  -- &
  -- &
  \multicolumn{1}{c|}{\textbf{98.94}} &
  \multicolumn{1}{c}{--} &
  \multicolumn{1}{c}{--} &
  \multicolumn{1}{c|}{\textbf{97.48}} &
  \multicolumn{1}{c}{--} &
  \multicolumn{1}{c}{--} &
  \textbf{81.17} \\
\tsupcon &
  98.01 &
  \textbf{98.14} &
  \multicolumn{1}{c|}{97.88} &
  97.61 &
  \textbf{97.88} &
  \multicolumn{1}{c|}{97.61} &
  \textbf{96.82} &
  96.42 &
  \multicolumn{1}{c|}{96.02} &
  89.39 &
  90.32 &
  \textbf{90.58} \\
\tsynproxy &
  \textbf{98.81} &
  97.61 &
  \multicolumn{1}{c|}{97.61} &
  \underline{\textbf{99.07}} &
  98.67 &
  \multicolumn{1}{c|}{95.49} &
  \textbf{97.75} &
  97.21 &
  \multicolumn{1}{c|}{95.76} &
  \textbf{96.42} &
  93.77 &
  36.87 \\ \bottomrule
\end{tabular}%
}
\label{tab:lrolis_lunar}
\end{table*}
% Please add the following required packages to your document preamble:
% \usepackage{booktabs}
% \usepackage{multirow}
% \usepackage{graphicx}
\begin{table*}[p]
\centering
\caption{Moon Lost-in-Space accuracy for models trained on Mars Crater, ablated by transformation type.}
\resizebox{0.825\textwidth}{!}{%
\begin{tabular}{@{}l|ccc|ccc|ccc|ccc@{}}
\toprule
\multicolumn{1}{c|}{\multirow{3}{*}{$\mathcal{L}_{\text{ML}}$}} &
  \multicolumn{3}{c|}{Illumination} &
  \multicolumn{3}{c|}{Translation} &
  \multicolumn{3}{c|}{Rotation} &
  \multicolumn{3}{c}{All} \\ \cmidrule(l){2-13} 
\multicolumn{1}{c|}{} &
  \begin{tabular}[c]{@{}c@{}}conv2d\\SE\end{tabular} &
  \begin{tabular}[c]{@{}c@{}}RIC\\CA\end{tabular} &
  \begin{tabular}[c]{@{}c@{}}MARs\\(Ours)\end{tabular} &
  \begin{tabular}[c]{@{}c@{}}conv2d\\SE\end{tabular} &
  \begin{tabular}[c]{@{}c@{}}RIC\\CA\end{tabular} &
  \begin{tabular}[c]{@{}c@{}}MARs\\(Ours)\end{tabular} &
  \begin{tabular}[c]{@{}c@{}}conv2d\\SE\end{tabular} &
  \begin{tabular}[c]{@{}c@{}}RIC\\CA\end{tabular} &
  \begin{tabular}[c]{@{}c@{}}MARs\\(Ours)\end{tabular} &
  \begin{tabular}[c]{@{}c@{}}conv2d\\SE\end{tabular} &
  \begin{tabular}[c]{@{}c@{}}RIC\\CA\end{tabular} &
  \begin{tabular}[c]{@{}c@{}}MARs\\(Ours)\end{tabular} \\ \midrule
\tcircle &
  96.42 &
  \textbf{98.28} &
  \textbf{98.28} &
  96.82 &
  96.42 &
  \underline{\textbf{98.81}} &
  95.49 &
  95.89 &
  \underline{\textbf{97.61}} &
  31.17 &
  69.36 &
  \textbf{88.73} \\
\tdrms &
  96.82 &
  \textbf{97.48} &
  96.68 &
  \textbf{97.21} &
  95.62 &
  96.82 &
  \textbf{95.89} &
  95.76 &
  95.49 &
  43.77 &
  \textbf{60.34} &
  46.68 \\
\tntxent &
  93.37 &
  95.62 &
  \textbf{96.95} &
  \textbf{91.38} &
  91.25 &
  91.11 &
  \textbf{94.83} &
  94.43 &
  93.90 &
  26.26 &
  45.09 &
  \textbf{48.01} \\
\tpnp &
  88.06 &
  92.18 &
  \textbf{95.36} &
  89.12 &
  88.06 &
  \textbf{89.26} &
  91.64 &
  \textbf{95.09} &
  92.97 &
  12.07 &
  \textbf{25.60} &
  24.27 \\
\tproxyanchor &
  96.82 &
  96.95 &
  \textbf{97.61} &
  97.48 &
  97.75 &
  \textbf{98.41} &
  96.29 &
  \textbf{96.68} &
  96.29 &
  41.25 &
  67.51 &
  \textbf{84.22} \\
\tproxynca &
  96.29 &
  \textbf{97.08} &
  96.55 &
  \textbf{97.88} &
  97.08 &
  97.75 &
  \textbf{96.68} &
  94.96 &
  96.42 &
  43.77 &
  57.29 &
  \textbf{70.29} \\
\tsubcenter &
  91.91 &
  95.49 &
  \textbf{96.29} &
  96.29 &
  \textbf{97.35} &
  96.55 &
  94.30 &
  \textbf{96.95} &
  96.68 &
  25.86 &
  47.75 &
  \textbf{54.64} \\
\tsupcon &
  90.98 &
  \textbf{97.48} &
  96.95 &
  89.52 &
  \textbf{93.50} &
  91.78 &
  92.44 &
  \textbf{95.76} &
  94.69 &
  16.98 &
  \textbf{68.30} &
  50.53 \\
\tsynproxy &
  97.75 &
  96.95 &
  \underline{\textbf{98.67}} &
  \textbf{98.67} &
  98.28 &
  98.54 &
  96.55 &
  96.02 &
  \textbf{97.08} &
  59.81 &
  67.37 &
  \underline{\textbf{89.52}} \\ \bottomrule
\end{tabular}%
}
\label{tab:lrolis_hirise}
\end{table*}

In this supplementary, we report additional evaluation including a transformation ablation and Mars Crater training results for Moon Navigation and Moon Lost-in-Space experiments. We also describe model execution times on spacecraft hardware, MARs per-block regularization curves, additional qualitative results, and Luna-1 dataset details. A summary of the results in this supplementary is as follows:
\begin{enumerate}
    \item When trained on Mars Crater, MARs model variants achieve the best accuracy for Moon Navigation in 4/9 $\ml$  (under all transformations).
    \item We discover a correlation between learned and architecture-encoded rotation equivariance with the amount of training data. With more training data (Moon Crater dataset), RIC CA diminishes recognition performance improving only 3/9 $\ml$ to conv2d SE in Moon Navigation considering all transformations. When trained with the sparser Mars Crater dataset, RIC CA improved 9/9 $\ml$. MARs (providing additional learned properties) improves 6/9 $\ml$ to RIC CA in Moon Navigation with Moon Crater and is highest overall in 4/9 $\ml$ losses (under all transformations).
    \item For Moon Lost-in-Space, MARs is highest overall considering all transformations in 4/9 and 6/9 $\ml$ trained with Moon Crater and Mars Crater respectively. 
    \item Qualitative examples demonstrate better attention alignment with MARs on post-training test images and per-epoch training evolutions.
\end{enumerate}

\section{Additional Results}
\subsection{Moon Navigation and Moon Lost-in-Space: Transformation Ablation and Mars Crater Training}

\autoref{tab:lrorec_lunar} and~\autoref{tab:lrorec_hirise} show Moon Navigation performance across all singular transformation types in $\mathcal{T}$ for model variants trained on Moon Crater and Mars Crater respectively. Our first observation is that RIC CA and MARs variants have impressive generalization to the lunar navigation sequence compared with baseline conv2d models, which fail to form representation spaces that transfer between the domains. This could be indicative of a correlation between the type of attention used as all $\ml$ losses benefit from spatial attention through CA layers. In many cases where RIC CA models only mildly improve performance MARs attention constraints lead to significant increases across each transform type (e.g., Circle, Proxy Anchor, Proxy Synthesis), where the opposite is also true in some aspects (e.g., ProxyNCA++). We speculate that this discrepancy in recognition can be attributed to how attention information resonates in the downstream embedding governed by certain $\ml$ losses that demonstrate a level of compatibility.

Secondly, we note that under rotations for Moon Navigation, RIC CA improves recognition performance only in 3/9 $\ml$ losses trained with Moon Crater data, which is in contrast to Mars Crater data where RIC CA improves all losses. This shows that with more training data available, fully explicit equivariant properties may be undesirable, and a \textit{partial} equivariance may be better suited instead. This supports the theory given in~\cite{leq3} which states that partial rotation equivariance is better suited to describe the relative pose of high-level features in the scene, whereas full-equivariance is better suited for low-level features. With more training data (Moon Crater), the learning framework is sufficient for many $\ml$ losses where low-level feature discrimination under rotations can be easily distinguished. With less training data (Mars Crater), explicit full rotation equivariance is needed for this low-level feature description. Following this evidence we observe that the extra learning-induced equivariant properties given by MARs lead to drastic performance increases in 4/9 $\ml$ terms (e.g., NTXent, PNP, Proxy Anchor, Subcenter ArcFace) across each transformation type.

\autoref{tab:lrolis_lunar} and~\autoref{tab:lrolis_hirise} show Moon Lost-in-Space accuracy for models trained on Moon and Mars data respectively across all singular transformation types in $\mathcal{T}$. We see similar behavior of the RIC CA models here, where performance over baseline conv2d variants is better with less training data (Mars) compared with more training data (Moon). This shows once again that the inclusion of explicit full rotation equivariant layers benefits scenarios with fewer training instances, which helps facilitate low-level feature discrimination under rotation transformations. This is further evident by observing the inclusion of MARs with Mars Crater data, where the extra implicit equivariance learning leads to greater improvements over RIC CA on 6/9 models considering all transformations. 

In general, the quantitative results show consistently high accuracy on all experiments using Proxy Anchor $\ml$ with RIC CA layers and MARs attention similarity constraints. This model-$\ml$ variant achieves either the highest or very competitive performance in all accuracy metrics studied and achieves state-of-the-art results on Moon Navigation experiments when observing all transformation types.

\subsection{MARs Regularization per ResNeXt Block}
% \begin{figure}[H]
% \centering
%   \includegraphics[width=\columnwidth]{figs/mars_block_loss.pdf}
%   \caption{MARs loss per ResNeXt block during the training of Proxy Anchor on Mars Crater.}
%   \label{fig:mars_block_loss}
% \end{figure}

\begin{wrapfigure}[12]{r}{0.5\textwidth}
\centering
  \includegraphics[width=0.49\textwidth]{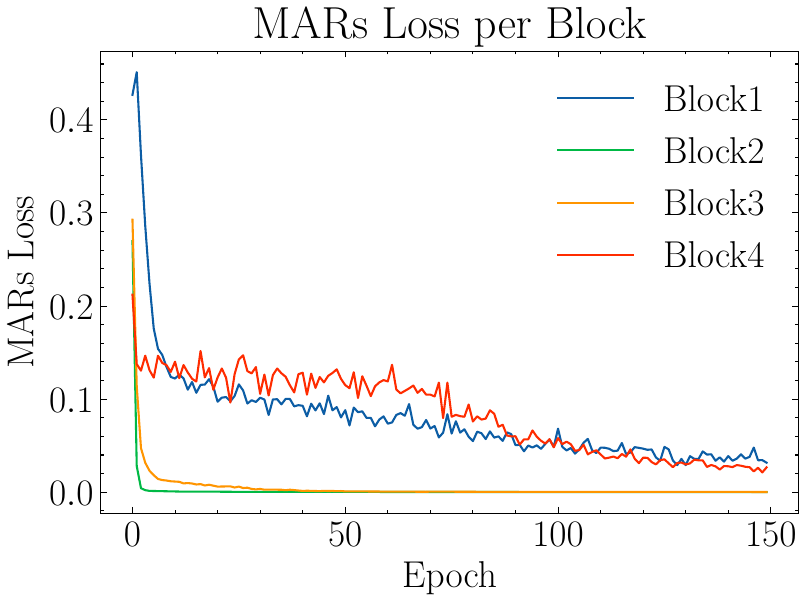}
  \caption{MARs regularization loss per ResNeXt block during the training of Proxy Anchor on Mars Crater.}
  \label{fig:mars_block_loss}
\end{wrapfigure}

\autoref{fig:mars_block_loss} shows independent loss value curves for MARs on each ResNeXt block in $f(\cdot)$ while training Proxy Anchor on Mars Crater. Of interesting note is the rapid convergence of MARs in the middle blocks (Block2 and Block3) compared with the outer blocks (Block1 and Block4). Block1 and Block4 have the highest dimensionality of spatial (32x32 in Block1) and channel (2048 in Block4) attention 
respectively, which may increase the difficulty of alignment. They are also the farthest (Block1) and closest (Block4) to the main learning $\ml$ term, indicating $\ml$'s immediate effect on Block4 but underpowered gradient updates to Block1. In contrast, the middle blocks learn attention alignment at a relatively independent rate potentially due to dimensionality differences or for being on the receiving end of immediate Block4 ``trickle down'' alignment. 

\subsection{Model Execution Times}
% Please add the following required packages to your document preamble:
% \usepackage{booktabs}
% \usepackage{graphicx}
\begin{table}[H]
\centering
\caption{Model execution times on a commodity laptop and representative spacecraft hardware platforms.}
\begin{tabular}{@{}c|ccccc@{}}
\toprule
Platform &
  Laptop CPU &
  Laptop GPU &
  \begin{tabular}[c]{@{}c@{}}Spacecraft\\ Hardware 1\end{tabular} &
  \begin{tabular}[c]{@{}c@{}}Spacecraft\\ Hardware 2\end{tabular} &
  \begin{tabular}[c]{@{}c@{}}Spacecraft\\ Hardware 3\end{tabular} \\ \midrule
Execution Time (ms) &
  37.31 &
  29.04 &
  12076 &
  846 &
  622 \\ \bottomrule
\end{tabular}
\label{tab:times}
\end{table}

% \autoref{tab:times} shows inference times of a single shot $f(\cdot)$ and $g(\cdot)$ on a commodity laptop and representative spacecraft hardware. 

To demonstrate in-situ feasibility, we examine the inference times of a single shot $f(\cdot)$ and $g(\cdot)$ on a commodity laptop and representative spacecraft hardware. The laptop is a Dell XPS, 12th Gen Intel i9-12900HK 2.5 GHz CPU, 64 GB RAM, NVIDIA RTX 3050 TI GPU. Our spacecraft hardware consists of NASA SpaceCube~\cite{sc3,sc3_mini} development boards with SpaceCube-LEARN (SC-LEARN,~\cite{sclearn}) or USB Google Coral Edge TPU deep learning accelerators. Spacecraft Hardware 1 is an AMD-Xilinx KCU105 evaluation board featuring a Kintex UltraScale FPGA and SC-LEARN. A MicroBlaze softcore processor is instantiated in the FPGA and runs a PetaLinux-based Linux operating system. This platform recently served as an emulator for the SCENIC mission onboard the International Space Station~\cite{scenic} and is a near-flight analog of the SpaceCube v3.0 Mini~\cite{sc3_mini}. Spacecraft Hardware 2 is nearly identical aside from a RISC-V-based Rocket Chip softcore processor, which runs a Yocto-based Linux operating system. Spacecraft Hardware 3 is a TUL PYNQ-Z2 evaluation board that features a Zynq-7020 SoC with a fixed-logic dual-core ARM Cortex-A9 processor and an Edge TPU USB Accelerator interfaced via USB 2.0. This setup is a development platform for the SpaceCube Mini-Z~\cite{sc3_mini}. Inference timings are shown in~\autoref{tab:times}, which showcase the utility of running MARs models onboard spacecraft flight computers at a higher rate than NFT (roughly 1.5 FPS best case vs NFT's 0.0083 FPS).

\subsection{Post-training Attention Visualizations}
\begin{figure*}[p]
     \centering
     \begin{subfigure}[]{0.38\textwidth}
         \centering
         \includegraphics[width=\textwidth]{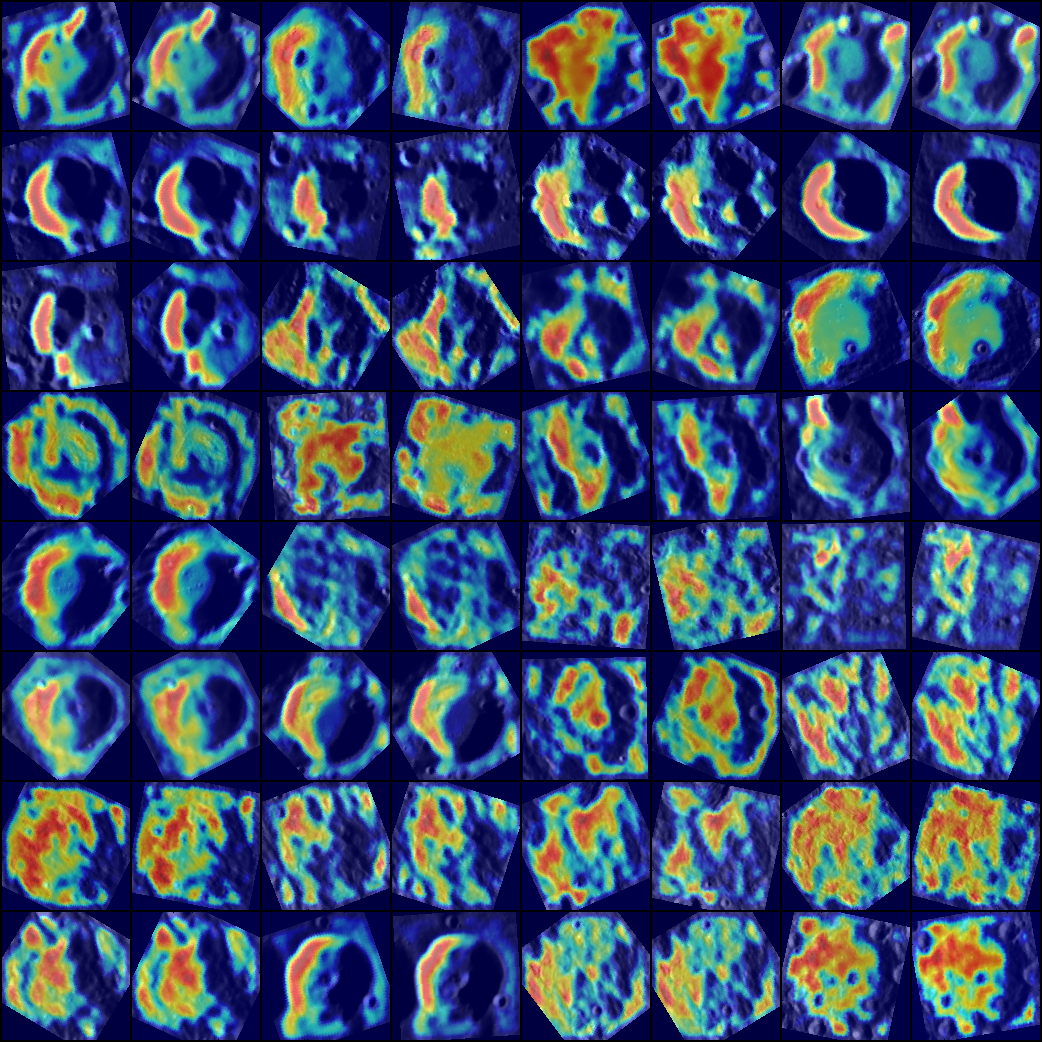}
         \caption{Moon Crater Block1, RIC CA}
     \end{subfigure}
     \begin{subfigure}[]{0.38\textwidth}
         \centering
         \includegraphics[width=\textwidth]{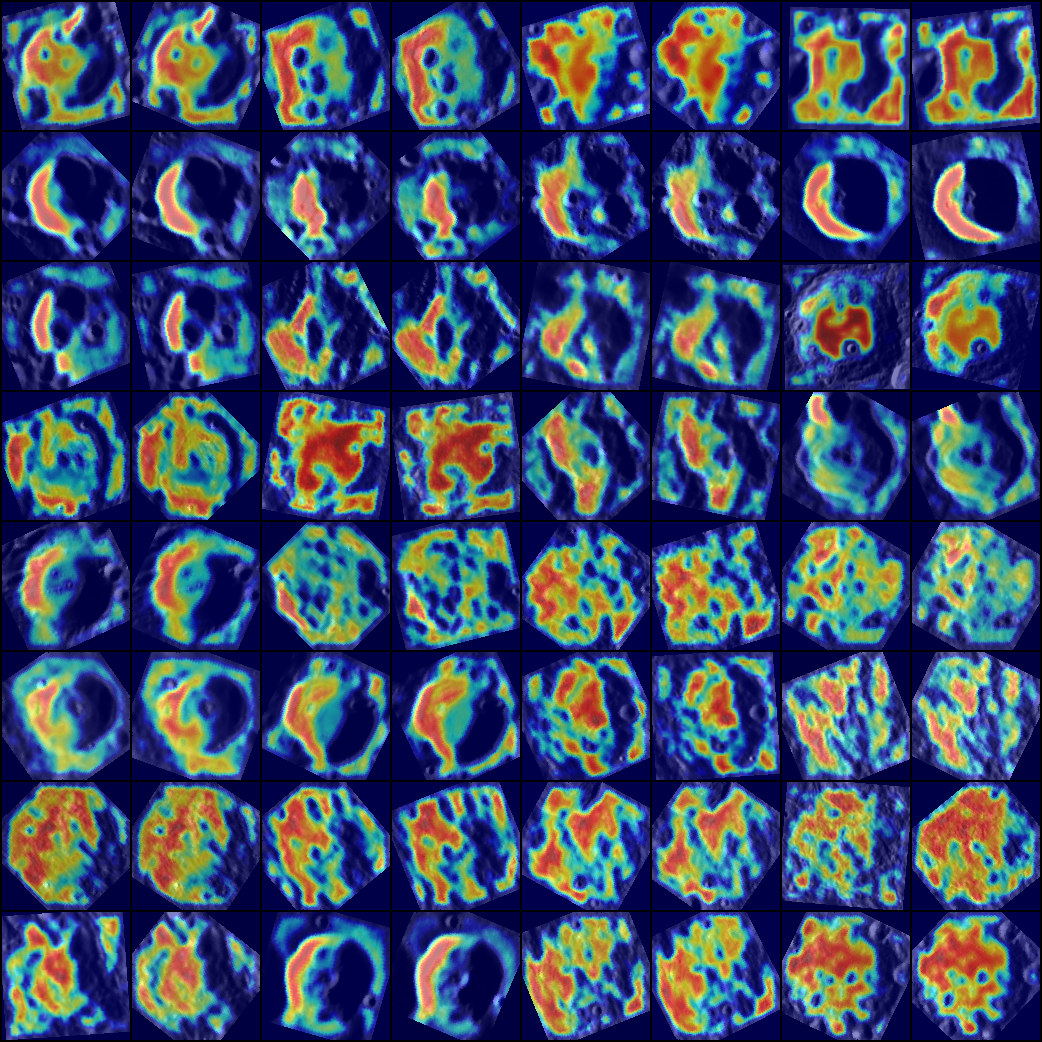}
         \caption{Moon Crater Block1, MARs}
     \end{subfigure}\\
     
     \begin{subfigure}[]{0.38\textwidth}
         \centering
         \includegraphics[width=\textwidth]{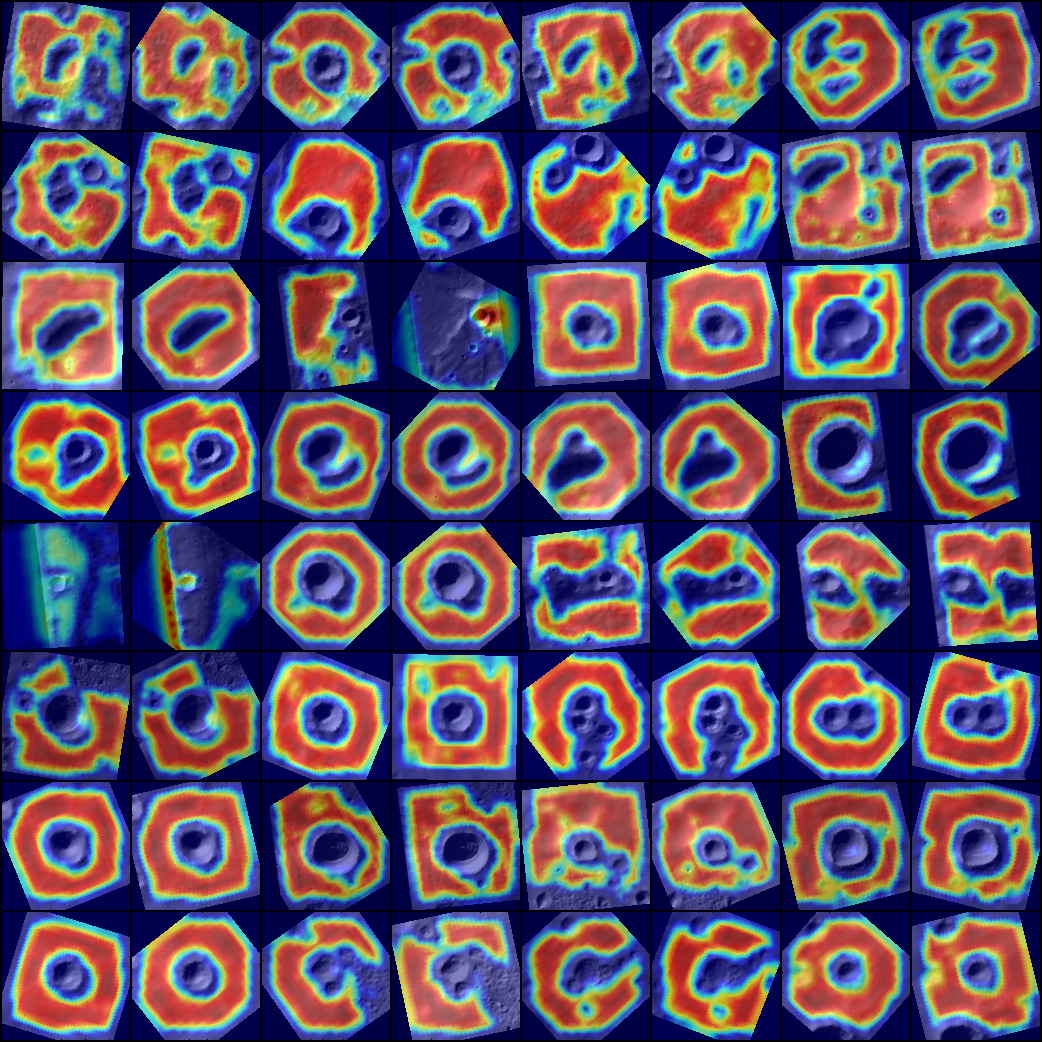}
         \caption{Mars Crater Block2, RIC CA}
     \end{subfigure}
     \begin{subfigure}[]{0.38\textwidth}
         \centering
         \includegraphics[width=\textwidth]{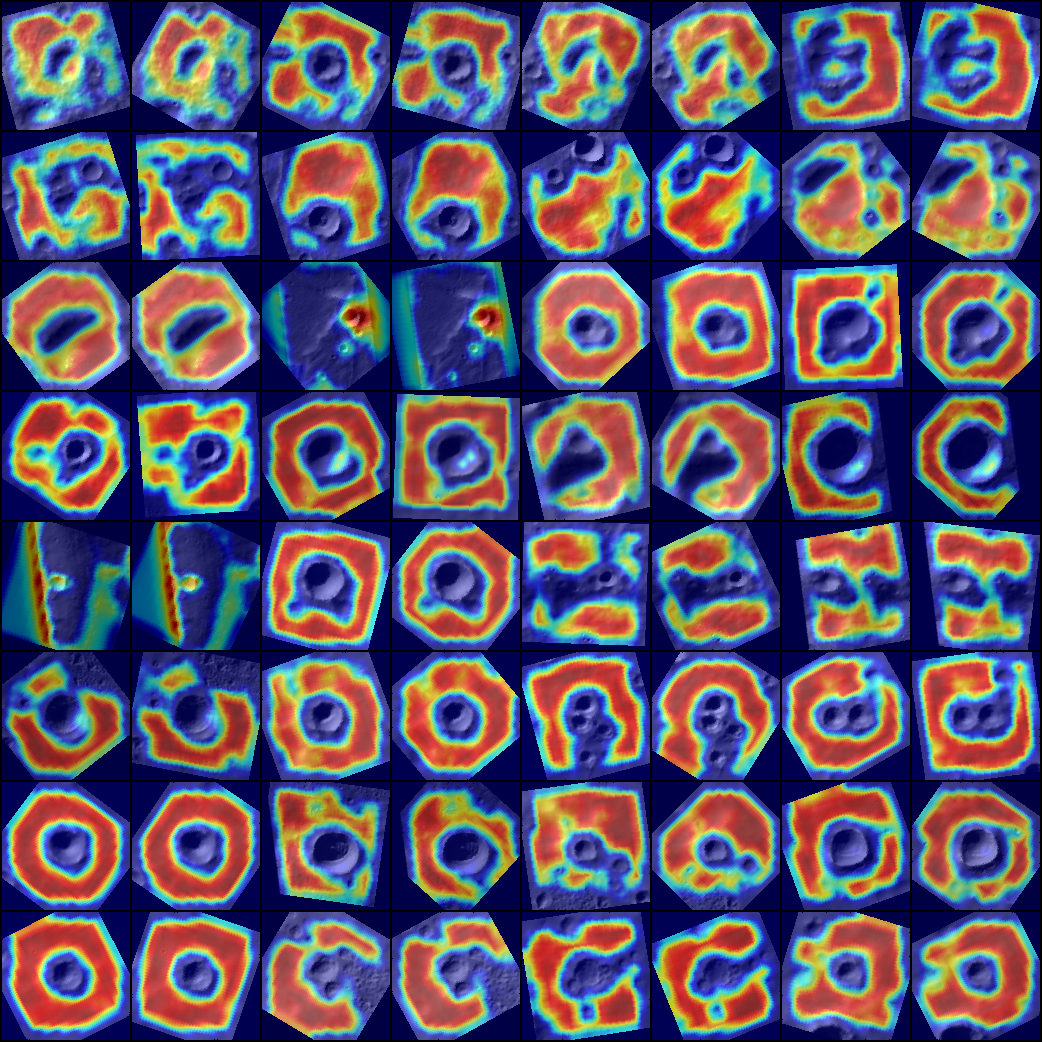}
         \caption{Mars Crater Block2, MARs}
     \end{subfigure}\\

     \begin{subfigure}[]{0.38\textwidth}
         \centering
         \includegraphics[width=\textwidth]{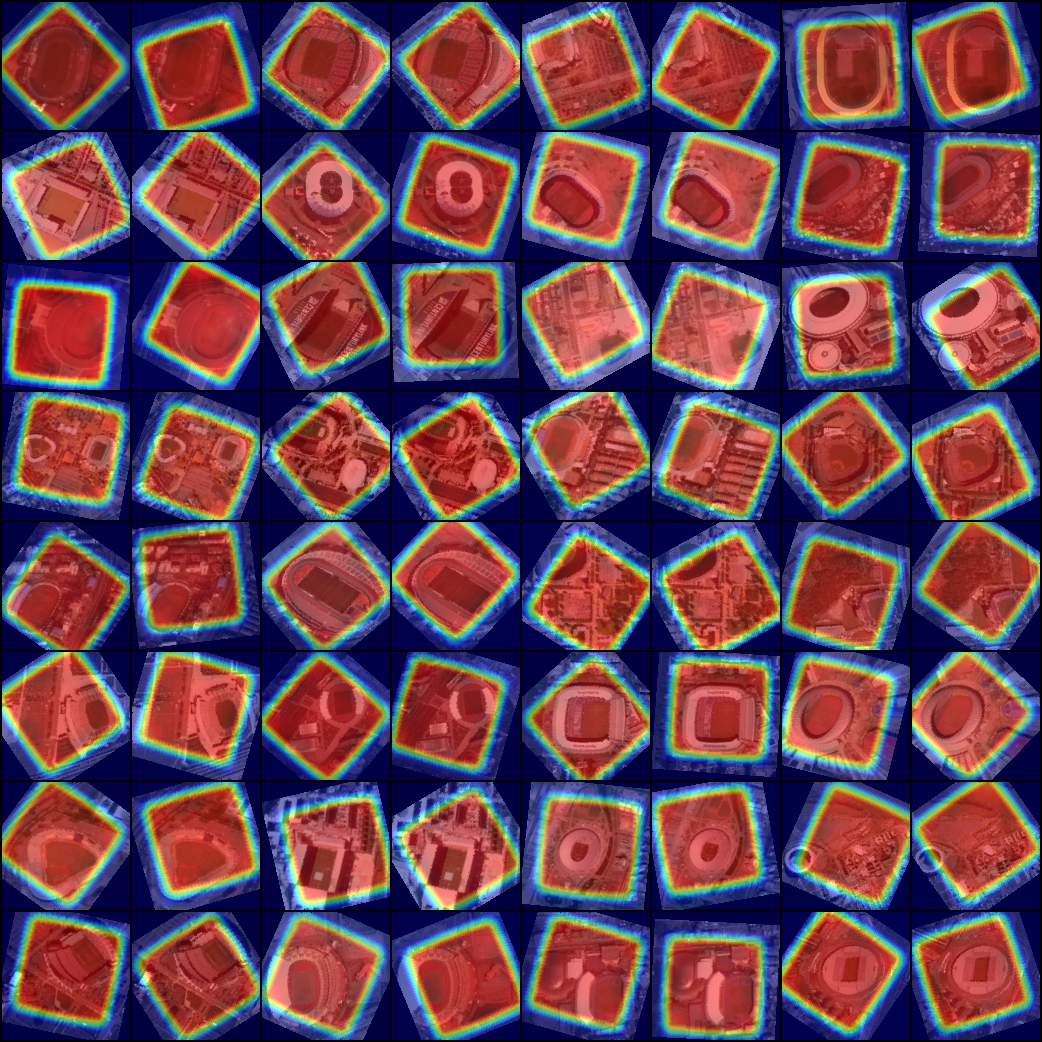}
         \caption{Earth Stadium Block3, RIC CA}
     \end{subfigure}
     \begin{subfigure}[]{0.38\textwidth}
         \centering
         \includegraphics[width=\textwidth]{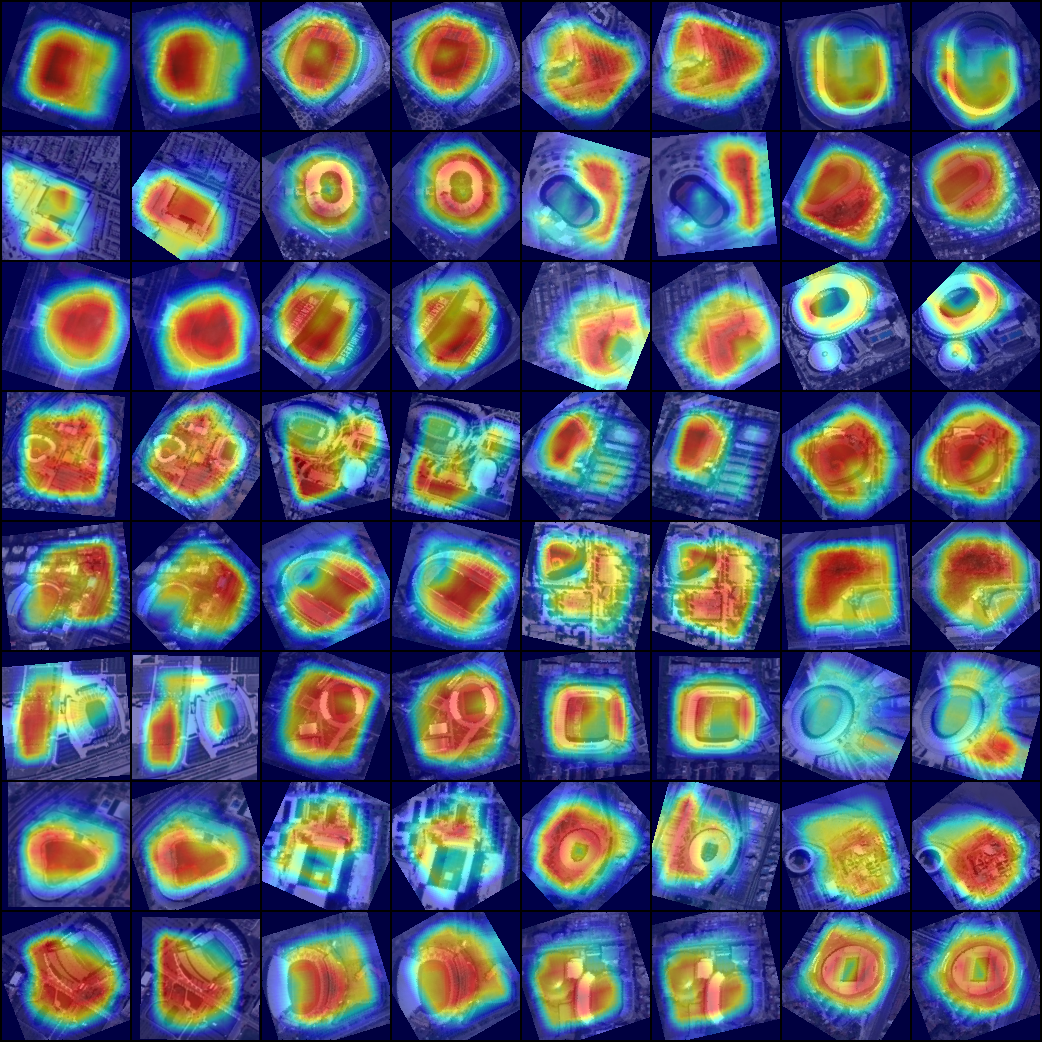}
         \caption{Earth Stadium Block3, MARs}
     \end{subfigure}

    \caption{Attention visualizations via the EigenCAM~\cite{eigencam} algorithm across Moon Crater, Mars Crater, and Earth Stadium datasets with $\ml=\text{NTXent, SupCon, and Proxy Synthesis}$ respectively. All MARs variants have $\gamma_{Sp} = 0.15, \gamma_{Ch} = 0.15$.}
    \label{fig:grids}
\end{figure*}

Additional post-training attention visualizations via EigenCAM~\cite{eigencam} for each dataset are shown in~\autoref{fig:grids}.

\subsection{Animations of Attention Evolution}
Animations showcasing multi-view attention evolution during training for each dataset are shown in~\autoref{fig:gifs}, visualized with EigenCAM~\cite{eigencam}. 

\section{Luna-1 Dataset Development Details}
\begin{figure*}[p]
    \centering
    \includegraphics[width=\textwidth]{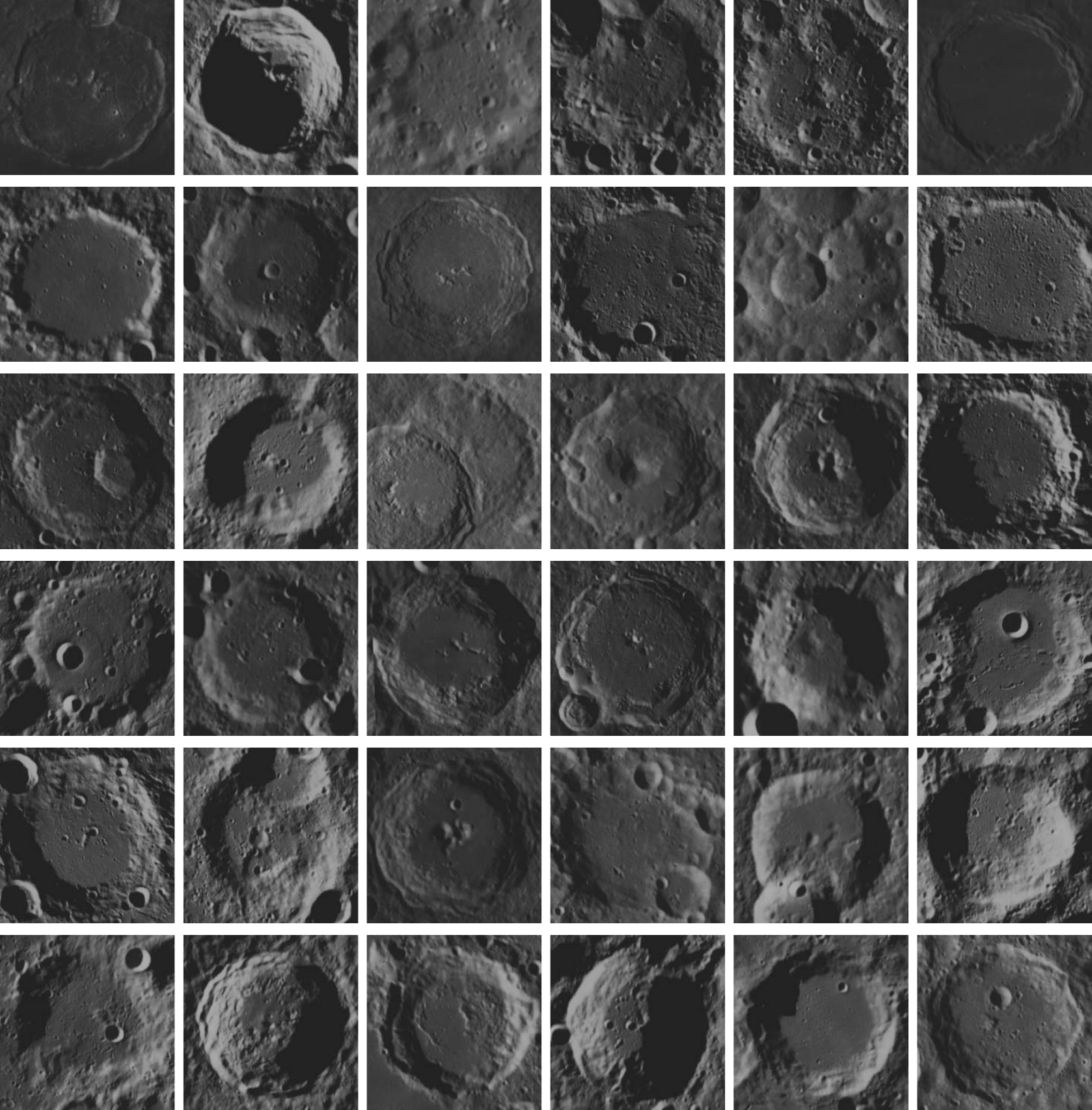}
    \caption{Example Moon Crater landmark renders from the Luna-1 dataset.}
    \label{fig:app_lunacraters}
\end{figure*}

Our photo-realistic dataset of a Moon environment is created using the Blender 3D software~\cite{blender}. Our base texture is a cylindrical projection mosaic at 64 pixels-per-degree~\cite{sim_texture} of resolution generated from LRO instrumentation, which gets displaced by a 64 pixels-per-degree elevation map~\cite{sim_dem} to add 3D topography. To facilitate the rendering of crater landmark images, we utilize the Lunar Orbiter Laser Altimeter (LOLA) Large Lunar Crater Catalog~\cite{crater_catalog} which provides roughly 5,000 crater locations in latitude/longitude/diameter format. For each crater in the catalog, we position the camera at its XYZ center point in the Blender environment and capture a frame that is the width of its diameter in pixels plus a margin of 25, and resize to 256x256 before rendering. The illumination source is kept tethered to the camera throughout the rendering process to ensure consistent lighting across each render. Any variations in illumination (including shadow cover) are artifacts of using true LRO imaging data, as the texture map reflects the real-world conditions when the spacecraft imaged each crater. Example crater renders are shown in~\autoref{fig:app_lunacraters}.

\begin{figure*}[p]
    \centering
    \includegraphics[width=\textwidth]{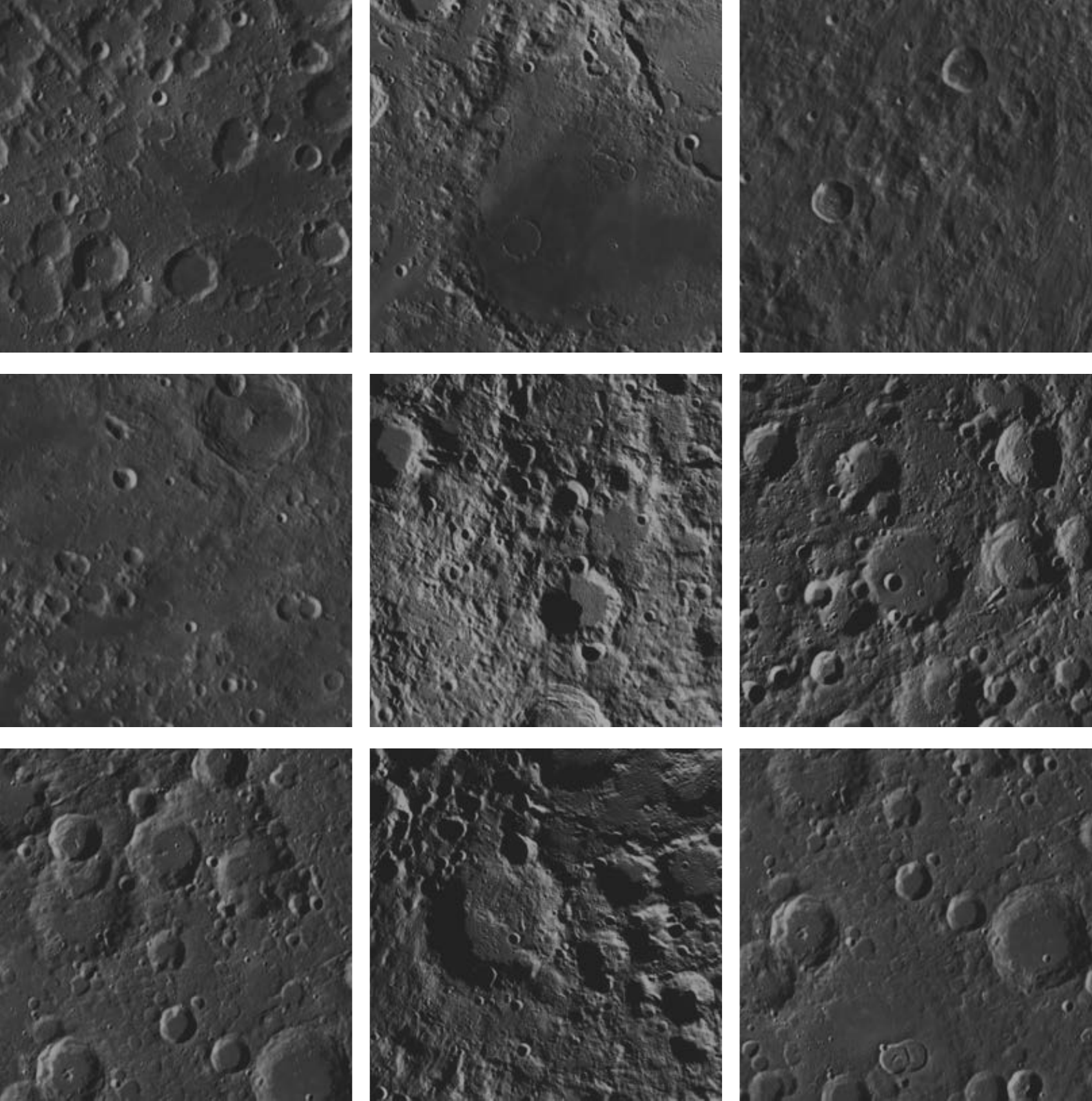}
    \caption{Example Moon navigation frame renders from the Luna-1 dataset.}
    \label{fig:app_lunagrid}
\end{figure*}

\begin{figure*}[p]
    \centering
    \includegraphics[width=\textwidth]{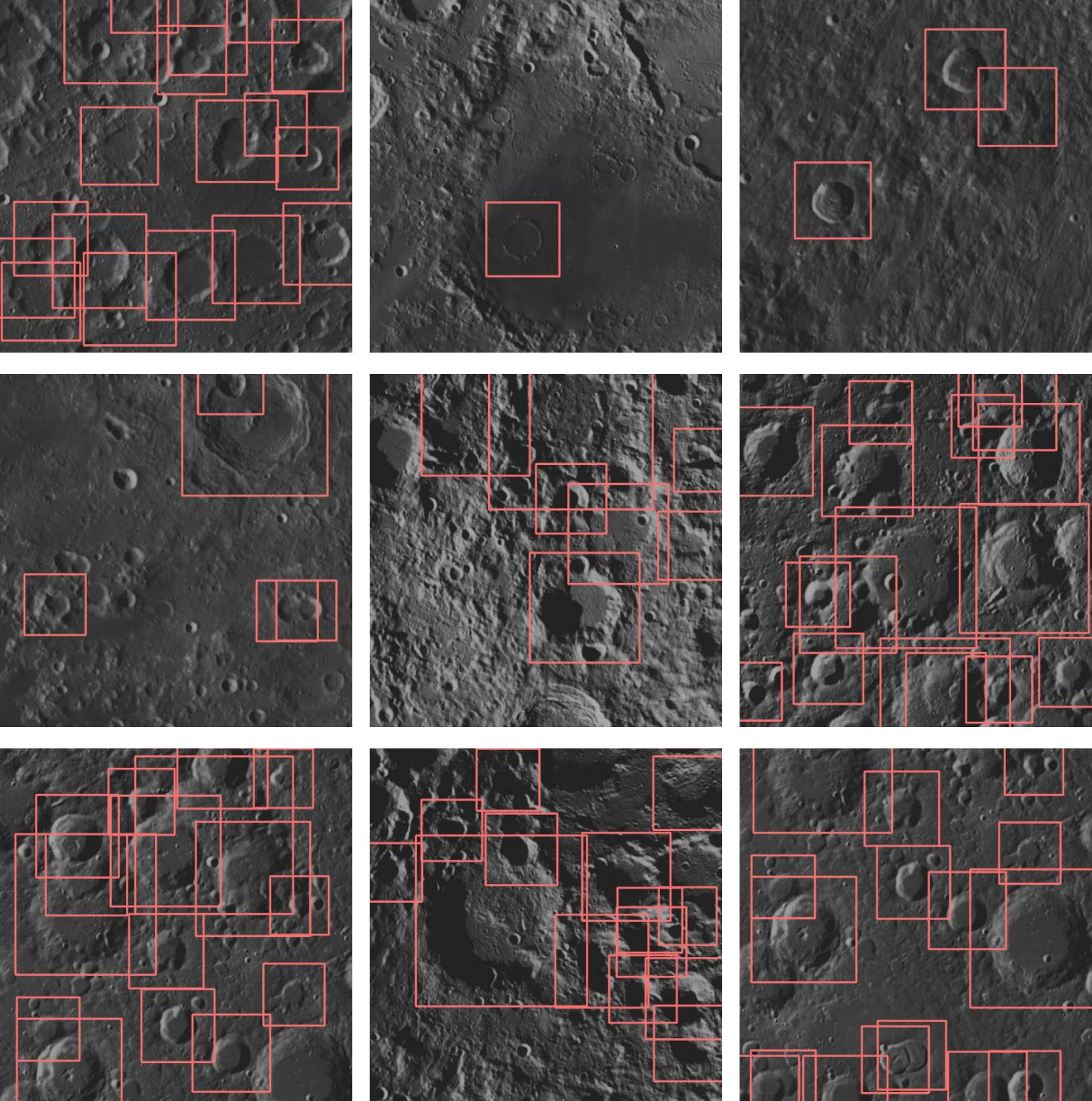}
    \caption{Example Moon navigation frame renders with visible crater annotations from the Luna-1 dataset.}
    \label{fig:app_lunabox}
\end{figure*}

To further enhance the dataset we pair the crater landmark renders with a real-world LRO spacecraft trajectory using historical spacecraft pose data gathered from NASA Navigation and Ancillary Information Facility (NAIF) SPICE kernels. We collect LRO pose information every 10 seconds throughout a 6-hour period starting from September 6th, 2022, which roughly correlates to 3 complete orbits of the spacecraft around the Moon. This length of trajectory ensures repeated visitation over regions and facilitates TRN and SLAM evaluation of place recognition and loop closure systems. We render image frames correlating to each collected LRO pose with its Z value (altitude) decreased as our Blender mesh of the Moon is a unit sphere. Additionally, we correlate crater locations from the catalog and provide bounding-box style annotations of visible craters within each frame. Unannotated and annotated examples of rendered LRO navigation frames are shown in~\autoref{fig:app_lunagrid} and~\autoref{fig:app_lunabox} respectively.
\clearpage

\begin{figure*}[p]
     \centering
     \begin{subfigure}[b]{0.49\textwidth}
         \centering
         \animategraphics[scale=0.6,controls,loop]{10}{figs/appendix/resisc_gif/seq-}{0}{149}
         \caption{Earth Stadium Block2, $\ml=\text{DR-MS}$, MARs $\gamma_{Sp} = 0.15, \gamma_{Ch} = 0.15$}
     \end{subfigure}
     \hfill
     \begin{subfigure}[b]{0.49\textwidth}
         \centering
         \animategraphics[scale=0.6,controls,loop]{10}{figs/appendix/hirise_gif/seq-}{0}{149}
         \caption{Mars Crater Block1, $\ml=\text{Proxy Anchor}$, MARs $\gamma_{Sp} = 0.3, \gamma_{Ch} = 0$}
     \end{subfigure}
     \\
     \vspace{1em}
     \begin{subfigure}[b]{\textwidth}
         \centering
         \animategraphics[scale=0.6,controls,loop]{10}{figs/appendix/lunar_gif/seq-}{0}{149}
         \caption{Moon Crater Block3, $\ml=\text{PNP}$, MARs $\gamma_{Sp} = 0.15, \gamma_{Ch} = 0.15$}
     \end{subfigure}
    \caption{Attention training animations of pose-normalized EigenCAMs~\cite{eigencam} from RIC CA (top) and MARs (bottom). These animations are viewable in Adobe Reader, KDE Okular, PDF-XChange, and Foxit Reader using the control bars.}
    \label{fig:gifs}
\end{figure*}

\clearpage

\clearpage

\clearpage

% ---- Bibliography ----
%
% BibTeX users should specify bibliography style 'splncs04'.
% References will then be sorted and formatted in the correct style.
%
\bibliographystyle{splncs04}
\bibliography{refs}